\documentclass[journal]{IEEEtran}

\usepackage{cite}

\usepackage{float}
\usepackage{enumerate}
\usepackage{graphicx} 
\usepackage{multirow} 
\usepackage{amsmath,amsfonts,amssymb}
\usepackage{makecell}
\usepackage{epstopdf} 
\usepackage{diagbox}
\usepackage{url}
\usepackage{xcolor}
\usepackage{threeparttable}
\usepackage[perpage,symbol]{footmisc}
\usepackage{cite} 
\usepackage{subfigure}
\setfnsymbol{wiley}

\usepackage{CJK}

\ifCLASSINFOpdf
\else
\fi
\begin{document}

\title{Adversarial Examples: Opportunities and Challenges}

\author{Jiliang~Zhang,~\IEEEmembership{Senior Member,~IEEE}, and Chen Li


\thanks{Manuscript received September 13, 2018; revised March 29, 2019, July 3, 2019, and July 15, 2019; accepted August 2, 2019. This work was supported in part by the National Natural Science Foundation of China under Grant 61874042 and Grant 61602107, in part by the Key Research and Development Program of Hunan Province under Grant 2019GK2082, in part by the Hu-Xiang Youth Talent Program under Grant 2018RS3041, in part by the Peng Cheng Laboratory Project of Guangdong Province PCL2018KP004, and in part by the Fundamental Research Funds for the Central Universities. (\textit{Corresponding author: Jiliang Zhang})}

\thanks{The authors are with the College of Computer Science and Electronic Engineering, Hunan University, Changsha 410082, China, and also with the Cyberspace Security Research Center, Peng Cheng Laboratory, Shenzhen 518000, China (e-mail: zhangjiliang@hnu.edu.cn).}

\thanks{Color versions of one or more of the figures in this paper are available online at http://ieeexplore.ieee.org.}

\thanks{Digital Object Identifier 10.1109/TNNLS.2019.2933524}

}


%
%

\markboth{IEEE Transactions on Neural Networks and Learning Systems,~2019}%
{Shell \MakeLowercase{\textit{et al.}}: Bare Demo of IEEEtran.cls for Journals}

%



\maketitle

\begin{abstract}
Deep neural networks (DNNs) have shown huge superiority over humans in image recognition, speech processing, autonomous vehicles and medical diagnosis. However, recent studies indicate that DNNs are vulnerable to adversarial examples (AEs), which are designed by attackers to fool deep learning models. Different from real examples, AEs can mislead the model to predict incorrect outputs while hardly be distinguished by human eyes, therefore threaten security-critical deep-learning applications. In recent years, the generation and defense of AEs have become a research hotspot in the field of artificial intelligence (AI) security. This article reviews the latest research progress of AEs. First, we introduce the concept, cause, characteristics and evaluation metrics of AEs, then give a survey on the state-of-the-art AE generation methods with the discussion of advantages and disadvantages. After that, we review the existing defenses and discuss their limitations. Finally, future research opportunities and challenges on AEs are prospected.

\end{abstract}


%
\IEEEpeerreviewmaketitle

\section{Introduction}
\IEEEPARstart{I}{n} recent years, deep neural networks (DNNs) have shown great advantages in autonomous vehicles, robotics, network security, image/speech recognition and natural language processing. For example, in 2017, an intelligent robot with the superior face recognition ability, named XiaoDu developed by Baidu, defeated a representative from the team of human’s strongest brain with the score of 3:2 \cite{Changjian2017}. On October 19, 2017, AlphaGo Zero released by the DeepMind team of Google shocked the world. Compared with the previous AlphaGo, AlphaGo Zero relies on reinforcement learning without any prior knowledge to grow chess skills and finally defeats every human competitor \cite{DemisHassabis2018}.

For artificial intelligence (AI) research, the United States received huge support from the government, such as the Federal Research Fund. In October 2016, the United States issued the projects of \emph{Preparing for the Future of Artificial Intelligence} and the \emph{National Artificial Intelligence Research and Development Strategic Plan}, which raised AI to the national strategic level and formulated ambitious blueprints \cite{Felten2016}, \cite{Biegel2016}. In 2017, China issued the \emph{New Generation Artificial Intelligence Development Plan}, which mentioned that the scale of the AI core industries would exceed 150 billion CNY by 2020, promoting the development of related industries to enlarge their scale to more than 1 trillion CNY. In the same year, AI was written into the \emph{nineteenth National Congress report}, which pushed the development of AI industries to a new height and filled the gap in the top-level strategy of AI development \cite{Graham2017}.

In the early stage of AI, people paid more attention to the basic theory and application research. With the rapid development of AI, security issues have attracted great attention. For example, at the Shenzhen Hi-tech Fair on November 16, 2016, a robot named Chubby suddenly broke down and hit the booth glass without any instructions and injured the pedestrian, which was the world's first robot injury incident \cite{ShailajaNeelakantan2016}. In July
2016, a crime-killing robot, Knightscope, manufactured by Silicon Valley Robotics, knocked down and injured a 16-month-old boy at the Silicon Valley shopping center \cite{McDonald2017}. At 10 p.m., March 22, 2018, an Uber autonomous test vehicle hit the 49-year-old woman named Elaine Herzberg who died after being sent to the hospital for invalid treatment in the suburbs of Tempe, Arizona. This is the first fatal autonomous vehicle accident in the world \cite{Siddiqui2018}.

\begin{figure}
\centerline{{\includegraphics[width=\linewidth]{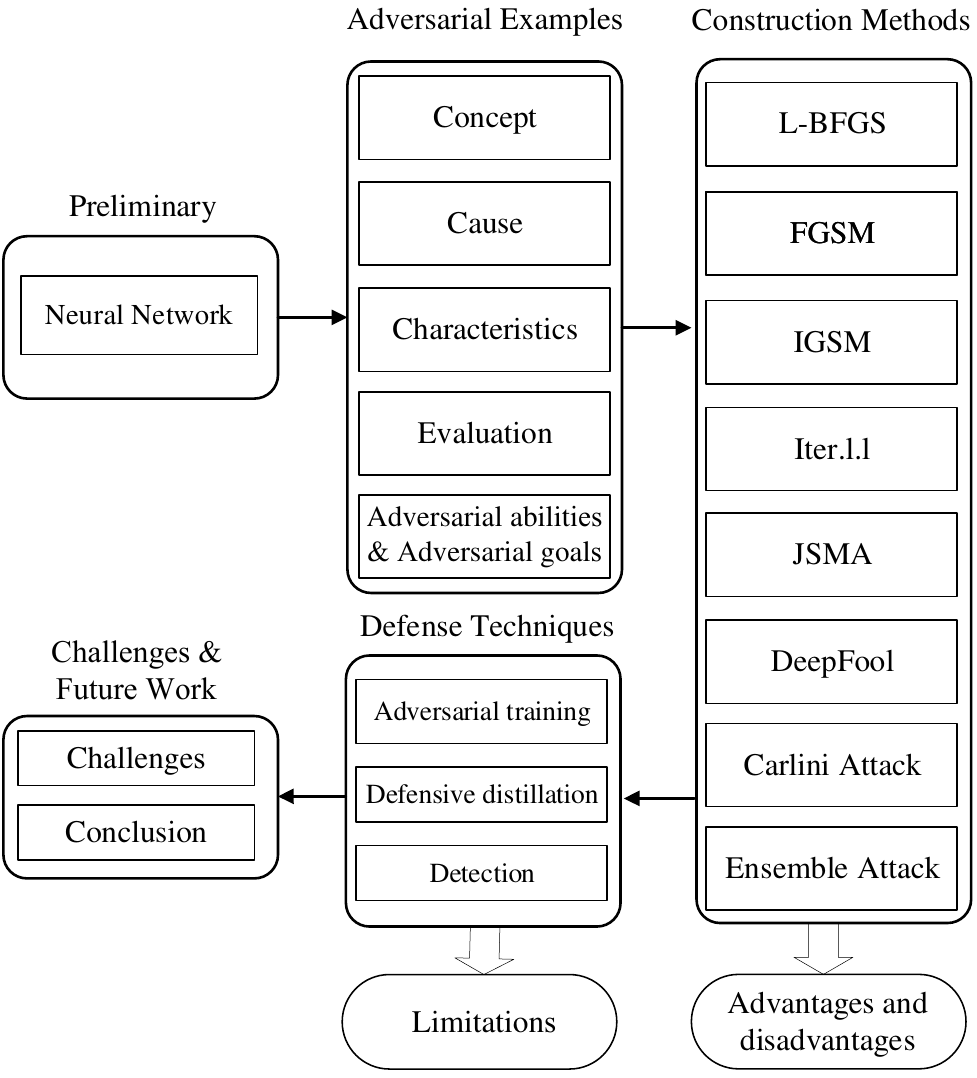}}}
\caption{The framework towards AEs for this review.}
\label{Fig1}
\end{figure}

In the past decade, various attacks on AI systems have emerged \cite{Biggio2018,Brundage2018a,Grosse2017d,Wittel,Biggio2012,Biggio,Biggio2013a,Abaid2017,Liu2019}. At the training stage, poisoning attacks \cite{Wittel,Biggio2012,Biggio} can damage the original probability distribution of training data by injecting malicious examples to reduce the prediction accuracy of the model. At the test or inference stage, evasion attacks \cite{Bulo2017,Grosse2017d,Biggio2013a,Abaid2017} can trick a target system by constructing a specific input example without changing the target machine learning (ML) system. In 2005, Lowd and Meek \cite{Lowd2005} proposed the concept of adversarial learning, in which an adversary conducts an attack to minimize a cost function. Under this framework, they proposed an algorithm to reverse engineer linear classifiers. In 2006, Barreno \emph{et al.} \cite{Barreno2006} presented a taxonomy of different types of attacks on ML systems. In order to mitigate poisoning attacks and evasion attacks, a lot of defenses have been proposed \cite{Biggio2018,Brundage2018a,Rubinstein2009,Domingos2004,Zhang2016}. In 2014, Szegedy \emph{et al.} \cite{Szegedy2013} proposed the concept of adversarial example (AE). By adding a slight perturbation to the input, the model misclassifies the AE with high confidence, while human eyes cannot recognize the difference. Even though different models have different architectures and training data, the same set of AEs can be used to attack related models. AEs have shown a huge threat to DNNs. For example, the classifier may misclassify an adversarial image of the stop traffic sign as a speed limit sign of 45 km/h, resulting in a serious traffic accident \cite{Garg2017}. In the image captioning system, an image is used as input to generate some captions to describe the image which is perturbed by attackers to generate some image-independent, completely opposite or even malicious captions \cite{Chen2017}. In the malware detection, the ML-based visualization malware detectors are vulnerable to AE attacks, where a malicious malware may be classified as a benign one by adding a slight perturbation on the transformed grayscale images \cite{Liu2019}.

\begin{figure}
\centerline{{\includegraphics[width=\linewidth]{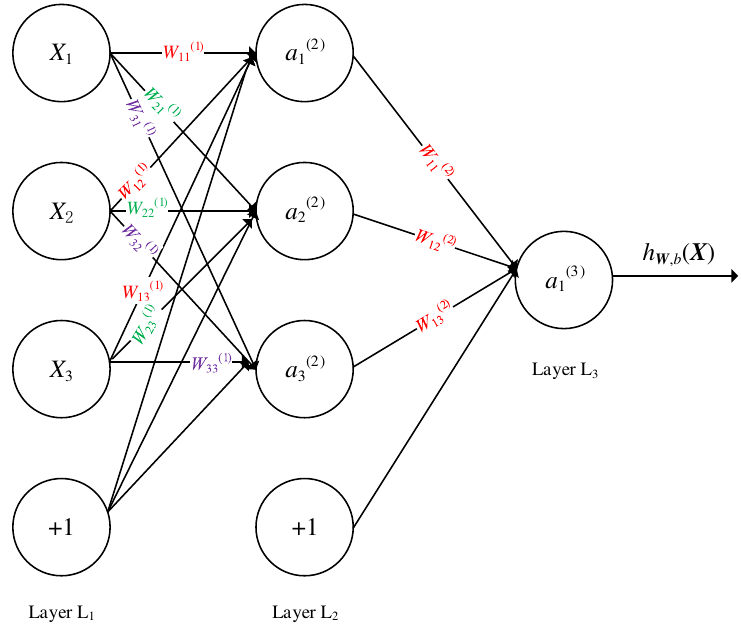}}}
\caption{A three-layer neural network model with the known inputs and weights.}
\label{Fig2}
\end{figure}

In recent years, many AE construction methods and defense techniques have been proposed. This survey elaborates on the related research and development status of AE on DNNs since it was proposed in \cite{Szegedy2013}. The overall framework is shown in Fig. \ref{Fig1}.

\section{Preliminaries}
\subsection{Neural Network}
Artificial neural network (ANN) simulates the human brain's nervous system to process complex information and consists of many interconnected neurons. Each neuron represents a specific output function called the activation function. The connection between two neurons represents the weight of the signal. The neural network connects many single neurons together by weights to simulate the human brain to process information.

As shown in Fig. \ref{Fig2}, a three-layer neural network is composed of an input layer $L_1$, a hidden layer $L_2$ and an output layer $L_3$, where the circle represents the neuron of the neural network; the circle labeled ``+1'' represents the bias unit; the circles labeled ``$X_1$'', ``$X_2$'', ``$X_3$'' are the inputs. Neurons in different layers are connected by weights \textbf{\emph{W}}. We use $a_{i}^{(l)}$ to represent the activation value (output value) of the \emph{i-th} unit in \emph{l-th} layer, when  \emph{l} = 1, $a_{i}^{(1)}$=$X_i$. With the given inputs and weights, the function output $h_{(\textbf{\emph{W}},b)}$(\textbf{\emph{X}}) can be calculated. The specific steps are as follows:
\begin{equation}
a_{1}^{(2)}=f(W_{11}^{(1)} X_1 + W_{12}^{(1)} X_2 + W_{13}^{(1)} X_3 + b_{11}^{(1)});
\end{equation}
\begin{equation}
a_{2}^{(2)}=f(W_{21}^{(1)} X_1 + W_{22}^{(1)} X_2 + W_{23}^{(1)} X_3 + b_{21}^{(1)});
\end{equation}
\begin{equation}
a_{3}^{(2)}=f(W_{31}^{(1)} X_1 + W_{32}^{(1)} X_2 + W_{33}^{(1)} X_3 + b_{31}^{(1)});
\end{equation}
\begin{equation}
a_{1}^{(3)}=h_{\textbf{\emph{W}},b} (\textbf{\emph{X}})=f(W_{11}^{(2)} a_{1}^{(2)} + W_{12}^{(2)} a_{2}^{(2)} + W_{13}^{(2)} a_{3}^{(2)} + b_{11}^{(2)}).
\end{equation}

The above-mentioned calculation process is called forward propagation (FP), which is a transfer process of input information through the hidden layer to the output layer.  The activation function rectified linear unit (ReLU): $f(X)=\max\{0,X\}$ is used to nonlinearize the neural network between different hidden layers. When the ML task is a binary classification, the final output layer uses the activation function sigmoid: $f(X)=1/(1+e^{(-X)})$. When the ML task is a multi-class problem, the final output layer uses the activation function softmax: $f(\textbf{\emph{X}})=({e^{X_k}}/({\sum_{i=1}^{N} e^{X_i}}))$, $k = 1, 2, ... N$. In the training process, the weights \textbf{\emph{W}} and the bias \emph{b} connecting the neurons in different layers are determined by back propagation.

\begin{figure}
\centerline{{\includegraphics[width=\linewidth]{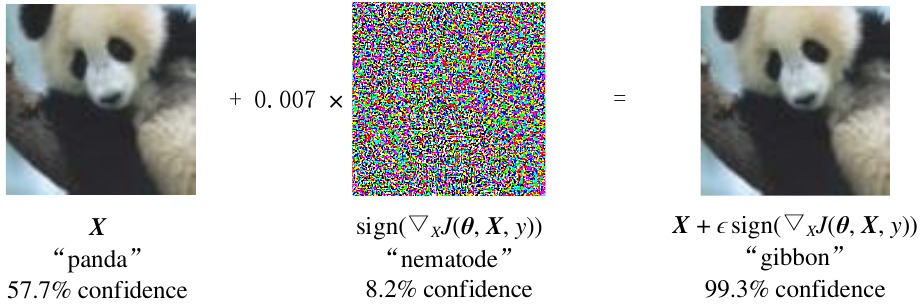}}}
\caption{Generating an AE with the FGSM, ``0.007'' corresponds to a small value $\epsilon$ that restricts the norm of the perturbation, ${\rm sign}(\nabla_X J(\boldsymbol{\theta},\textbf{\emph{X}},y))$ represents an imperceptible perturbation \cite{Goodfellow2014}.}
\label{Fig4}
\end{figure}

Neural networks belong to a cross-disciplinary research field combining computer, probability, statistics, and brain science. They focus on how to enable computers to simulate and implement human learning behaviors, so as to achieve better automatic knowledge acquisition. However, recent studies show that neural networks are particularly vulnerable to AEs which are generated by adding small perturbations to the inputs. In what follows, we will discuss the AEs in detail.

\begin{figure*}
\centerline{{\includegraphics[width=15cm,height=4cm]{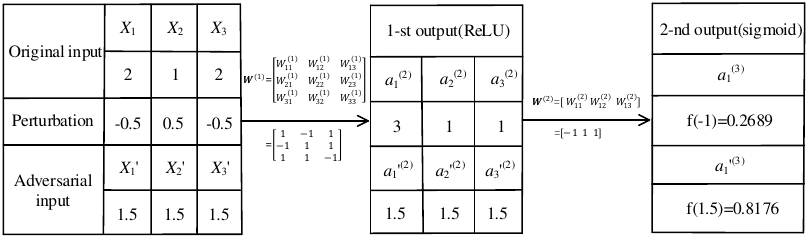}}}
\caption{The perturbation to the original inputs for a three-layer neural network (Fig. \ref{Fig2}). The value of the original inputs $X_1, X_2, X_3$ and the weights $\textbf{\emph{W}}^{(1)}, \textbf{\emph{W}}^{(2)}$ are initialized randomly.}
\label{Fig3}
\end{figure*}

\section{Adversarial Examples}
In 2014, Szegedy \emph{et al.} \cite{Szegedy2013} proposed AEs to fool DNNs. Adding a subtle perturbation to the input of the neural network will produce an error output with high confidence, while human eyes cannot recognize the difference. Suppose that there are a ML model $M$ and an original example $C$ which can be correctly classified by the model, i.e., $M(C)=y_{\rm true}$, where $y_{\rm true}$ is the true label of $C$. However, it is possible to construct an AE $C^{'}$ which is perceptually indistinguishable from $C$ but is classified incorrectly, i.e., $M(C^{'})\neq y_{\rm true}$ \cite{Szegedy2013}. A typical example is shown in Fig. \ref{Fig4}, the model considers the original image to be a ``panda'' (57.7$\%$). After adding a slight perturbation to the original image, it is classified as a ``gibbon'' by the same model with 99.3$\%$ confidence, while the human eyes completely cannot distinguish the differences between the original image and the adversarial image \cite{Goodfellow2014}.

In order to facilitate the reader to understand AEs intuitively, we use the neural network model in Fig. \ref{Fig2} as an example to show the change of the outputs by perturbing the inputs. As shown in Fig. \ref{Fig3}, $\textbf{\emph{W}}^{(1)}$ and $\textbf{\emph{W}}^{(2)}$ are the weight matrices. After adding a small perturbation sign (0.5) to the original inputs, the adversarial inputs $X_{1}^{'}$, $X_{2}^{'}$, $X_{3}^{'}$ are equal to 1.5. Then, through the first layer's weight matrix $\textbf{\emph{W}}^{(1)}$ and the transform operation of the activation function ReLU, the output values $a_{1}^{'(2)}$, $a_{2}^{'(2)}$, $a_{3}^{'(2)}$ are equal to 1.5. Finally, after passing the second layer's weight matrix $\textbf{\emph{W}}^{(2)}$ and the activation function sigmoid transform operation, the probability of the output is changed from 0.2689 to 0.8176, which makes the model misclassify the image with high confidence. With the increase of the model depth, the probability of the output changes more obviously.

\subsection{Cause of Adversarial Examples}
AE is a serious vulnerability in deep learning systems and cannot be ignored in security-critical AI applications. However, in current research, there are no well-recognized explanations on why the AEs can be constructed. Analyzing the cause of AEs can help researchers to fix the vulnerability effectively. The reason may be overfitting or insufficient regularization of the model which leads to insufficient generalization ability that learning models predict unknown data. However, by adding perturbations to a regularized model, Goodfellow \emph{et al.} \cite{Goodfellow2014} found that the effectiveness against AEs was not improved significantly. Other researchers \cite{Philipp2018} suspected that AEs arose from extreme nonlinearity of DNNs. However, if the input dimensions of a linear model are high enough, AEs can also be constructed successfully with high confidence by adding small perturbations to the inputs.

Goodfellow \emph{et al.} \cite{Goodfellow2014} believed that the reason for generating AEs is the linear behavior in high dimensional space. In the high dimensional linear classifier, each individual input feature is normalized. For one dimension of each input, small perturbations will not change the overall prediction of the classifier. However, small perturbations to all dimensions of the inputs will lead to an effective change of the output.

\begin{figure}
\centerline{{\includegraphics[width=\linewidth]{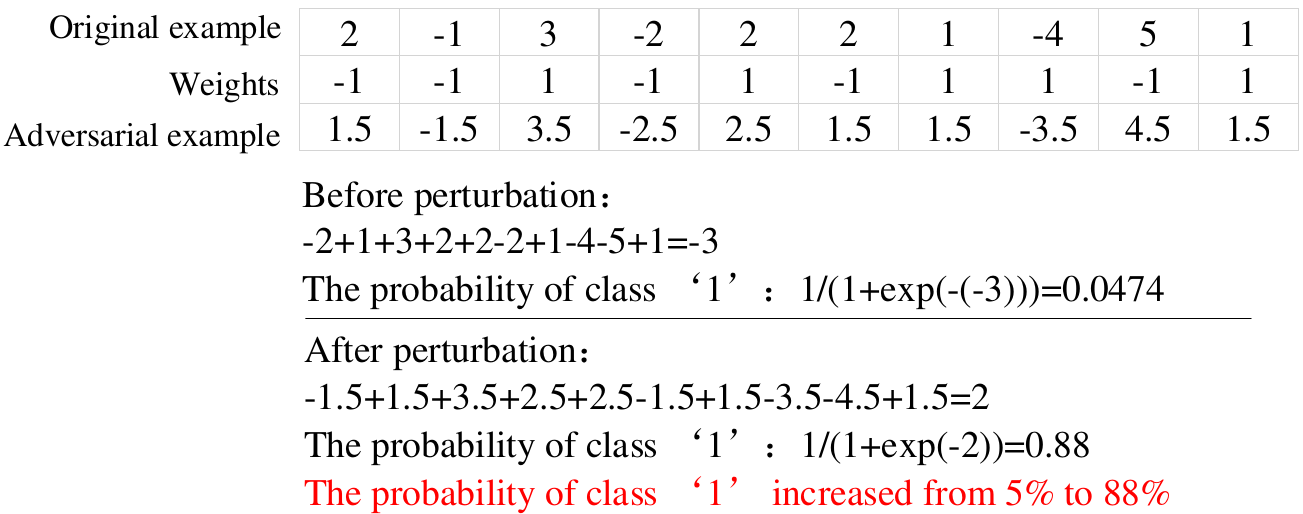}}}
\caption{The probability of class `1' before and after the perturbation.}
\label{Fig5}
\end{figure}

As shown in Fig. \ref{Fig5}, the score of class '1' is improved from 5$\%$ to 88$\%$ by adding or subtracting 0.5 to each dimension of the original example \textbf{\emph{X}} in a particular direction. It demonstrates that linear models are vulnerable to AEs and refutes the hypothesis that the existence of AEs is due to the high nonlinearization of the model. Therefore, the existence of high-dimensional linear space may be the cause of AEs.

\subsection{Characteristics of Adversarial Examples}
In general, AEs have three characteristics:

{\bfseries Transferability.} AEs are not limited to attack a specific neural network. It is unnecessary to obtain architecture and parameters of the model when constructing AEs, as long as the model is trained to perform the same task. AEs generated from one model $M_1$ can fool a different model $M_2$ with a similar probability. Therefore, an attacker can use AEs to attack the models that perform the same task, which means that an attacker can construct AEs in the known ML model and then attack related unknown models \cite{Papernot2016a}.

{\bfseries Regularization effect.} Adversarial training \cite{Goodfellow2014} can reveal the defects of models and improve the robustness. However, compared to other regularization methods, the cost of constructing a large number of AEs for adversarial training is expensive. Unless researchers can find shortcuts for constructing AEs in the future, they are more likely to use dropout \cite{Hinton2014} or weight decay ($L_2$  regularization).

{\bfseries Adversarial instability.} In the physical world, it is easy to lose the adversarial ability for AEs after physical transformations such as translation, rotation, and lighting. In this case, AEs will be correctly classified by the model. This instability characteristic challenges attackers to construct robust AEs and creates the difficulty of deploying AEs in the real world.

\begin{figure}
\centerline{{\includegraphics[width=6.5cm]{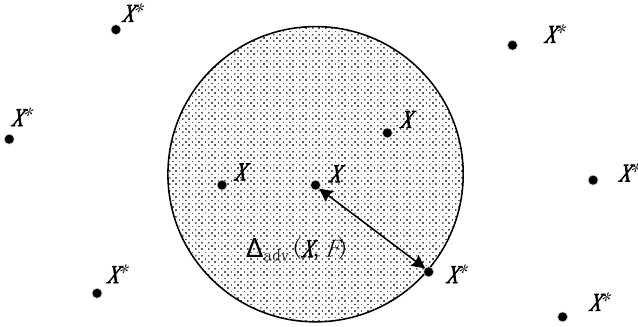}}}
\caption{\textbf{Visualizing the metric of robustness:} This 2D representation illustrates the metric as the radius of the disc at the original example \textbf{\emph{X}} and going through the closest AE $\textbf{\emph{X}}^{*}$ constructed from \textbf{\emph{X}} \cite{Papernot2016}.}
\label{Fig6}
\end{figure}

\subsection{Evaluation Metrics}
\subsubsection{Success Rate}

When constructing AEs, the success rate is the most direct and effective evaluation criterion. In general, the success rate to generate AEs is inversely proportional to the magnitude of perturbations. For example, the fast gradient sign method \cite{Goodfellow2014} requires a large perturbation and is prone to label leaking \cite{Kurakin2016a} that the model correctly classifies the AE generated with the true label and misclassifies the AE created with the false label. Therefore, the success rate is much lower than the iterative method \cite{Kurakin2016} with the lower perturbation and the Jacobian-based saliency map attack method \cite{Papernot2016b} with the specific perturbation. Usually, it is difficult to construct AEs with 100$\%$ success rate.

\subsubsection{Robustness}

The robustness of ML models is related to the classification accuracy \cite{Su}, \cite{Rozsa2017}. Better ML models are less vulnerable to AEs. Robustness is a metric to evaluate the resilience of DNNs to AEs. In general, a robust DNN model has two features \cite{Carlini2016} \cite{Fawzi2018}:
\begin{itemize}
\item The model has high accuracy both inside and outside of the dataset;
\end{itemize}

\begin{itemize}
\item The classifier of a smoothing model can classify inputs consistently near a given example.
\end{itemize}

We first define the robustness of classifiers $f(\textbf{\emph{X}})$ to adversarial perturbations in the input space $\mathbb{R}^d $. Given an input $\textbf{\emph{X}}$ $\in$ $\mathbb{R}^d $ from $\mu$ (the probability measure of the data points that we wish to classify), $\Delta_{\rm adv}(\textbf{\emph{X}},F)$ is denoted as the norm of the smallest perturbation to make models misclassified.
\begin{equation}
\begin{split}
\Delta_{\rm adv}(\textbf{\emph{X}},F)=\min \limits_{\textbf{\emph{r}} \in \mathbb{R}^d}\ ||\textbf{\emph{r}}||_2 ,\quad \\
{\rm s.t.}\ F(\textbf{\emph{X}})F(\textbf{\emph{X}}+\textbf{\emph{r}})\leq 0,
\end{split}
\end{equation}
where the perturbation $\textbf{\emph{r}}$ aims to flip the label of $\textbf{\emph{X}}$, corresponding to the minimal distance from $\textbf{\emph{X}}$ to the decision boundary of the classifier.

The robustness of a DNN model $F$ to adversarial perturbations is defined as the average of $\Delta_{\rm adv}(\textbf{\emph{X}},F)$ over all $\textbf{\emph{X}}$, and the corresponding expression is
\begin{equation}
\rho_{\rm adv}(F)=E_\mu[\Delta_{\rm adv}(\textbf{\emph{X}},F)].
\end{equation}

As shown in Fig. \ref{Fig6}, the outputs of the classifier are constant inside the circle with a radius of $\Delta_{\rm adv}(\textbf{\emph{X}},F)$. However, the classified results of all samples $\textbf{\emph{X}}^{*}$ outside the circle are different from \textbf{\emph{X}}. Therefore, the magnitude of the perturbation $\Delta_{\rm adv}(\textbf{\emph{X}},F)$ is proportional to the robustness of the model, i.e., the higher the minimum perturbation needed to misclassify the example, the stronger the robustness of the DNN is.

\subsubsection{Transferability}

AEs generated for one ML model can be used to misclassify another model even if both models have different architectures and training data. This property is called transferability. AEs can be transferred among different models because a contiguous subspace with a large dimension in the adversarial space is shared among different models \cite{Tramer2017}. This transferability provides a tremendous advantage for AEs because attackers only need to train alternative models to construct AEs and deploy them to attack the target model.

The transferability of AEs can be measured by the transfer rate, i.e., the ratio of the number of transferred AEs to the total number of AEs constructed by the original model. In the non-targeted attack, the percentage of the number of AEs generated by one model that are correctly classified by another model is used to measure the non-targeted transferability. It is called the accuracy rate. A lower accuracy rate means a better non-targeted transfer rate. In the targeted attack, the percentage of the AEs generated by one model that can be classified by another model as the target label is used to measure the targeted transferability. It is referred to as the matching rate, and a higher matching rate means a better targeted transfer rate \cite{Liu2016}.

The transfer rate of AEs depends on two factors. One is the model-related parameters, including the model architecture, model capacity, and test accuracy. The transfer rate of AEs is high among models with similar architecture, low model capacity (the number of model parameters) and high test accuracy \cite{Wu2018}. Another factor is the magnitude of the adversarial perturbation. Within a certain perturbation range, the transfer rate of AEs is proportional to the magnitude of adversarial perturbations, i.e., the greater perturbations to the original example, the higher transfer rate of the constructed AE. The minimum perturbation required for different methods of constructing AEs is different.

\subsubsection{Perturbations}
Too small perturbations on the original examples are difficult to construct AEs, while too large perturbations are easily distinguished by human eyes. Therefore, perturbations need to achieve a balance between constructing AEs and the human visual system. For example, it is difficult to control the perturbation for FGSM \cite{Goodfellow2014} which incurs label leaking easily. To address the issue, Kurakin \emph{et al.} proposed an optimized FGSM based on the iterative method \cite{Kurakin2016} which can control the perturbation within a threshold range. Hence, the success rate of constructing AEs is improved significantly. However, the transfer rate of such AEs is low. Later on, a saliency map-based method \cite{Papernot2016b} is proposed to improve the transfer rate. The key steps include: 1) \emph{direction sensitivity estimation}: evaluate the sensitivity of each class for each input feature; 2) \emph{perturbation selection}: use the sensitivity information to select a minimum perturbation $\boldsymbol{\delta}$ among the input dimension which is most likely to misclassify the model.


In general, $L_2$-norm is used to measure the perturbation of the AE and is defined as
\begin{equation}
d(\textbf{\emph{X}}^{'},\textbf{\emph{X}})={||\textbf{\emph{X}}^{'}-\textbf{\emph{X}}||}_2=\sqrt{\sum_{i=1}^n(X_i^{'}-X_i)^2},
\end{equation}
where the n-dimensional vectors $\textbf{\emph{X}}$ and $\textbf{\emph{X}}^{'}$ represent the original example and AE respectively, $d(\textbf{\emph{X}}^{'},\textbf{\emph{X}})$ is the distance metric between \textbf{\emph{X}} and $\textbf{\emph{X}}^{'}$. It shows that the larger $d(\textbf{\emph{X}}^{'},\textbf{\emph{X}})$ to the original example, the greater perturbations needed to construct AEs.

\subsubsection{Perceptual adversarial similarity score}

 AEs are visually recognized by humans as correct classes while being misclassified by the models. Since the human visual system is sensitive to structural changes, Wang \emph{et al.} \cite{Wang2004} proposed the structural similarity (SSIM) index as a metric to measure the similarity between two images. Luminance and contrast associated with the object structure are defined as the structure information of the image.

\begin{figure}
\centerline{{\includegraphics[width=\linewidth]{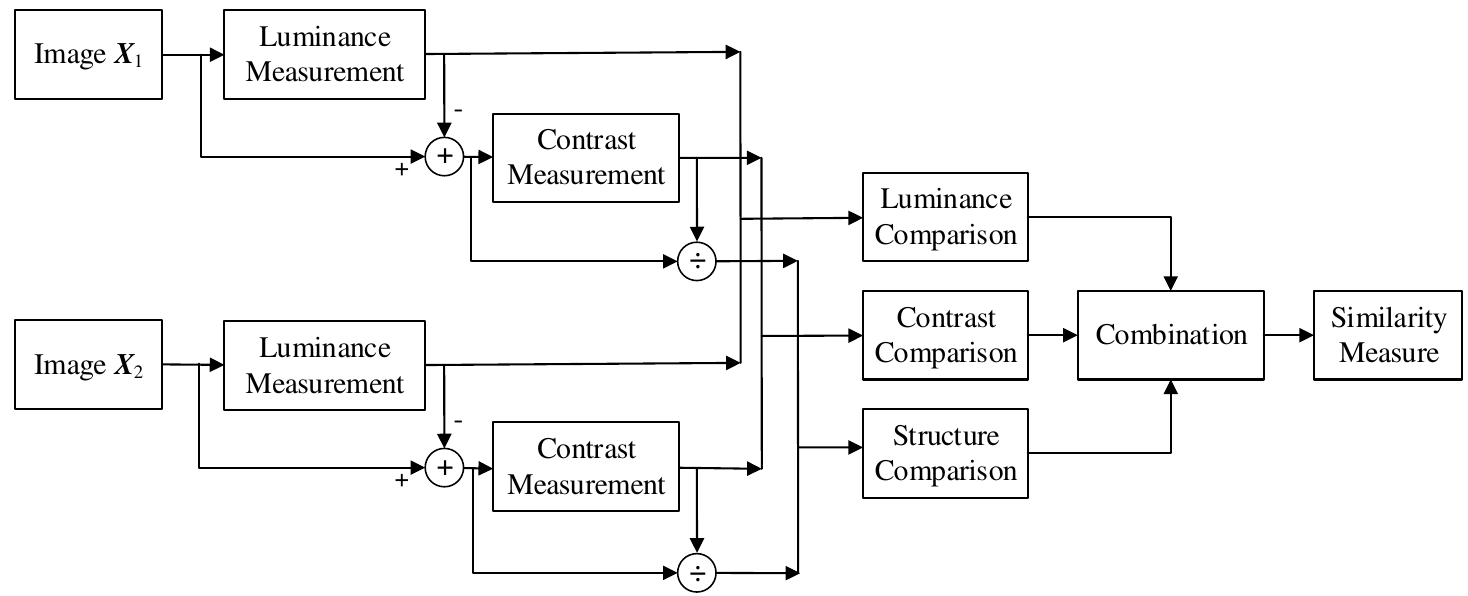}}}
\caption{The structural similarity (SSIM) measurement system \cite{Wang2004}.}
\label{Fig7}
\end{figure}

The structure of the SSIM measurement system is shown in Fig. \ref{Fig7}. For two different aligned images $\textbf{\emph{X}}_1$  and $\textbf{\emph{X}}_2$, the SSIM measurement system consists of three comparisons: luminance, contrast, and structure. First, the luminance of each image is compared. Second, the standard deviation (the square root of variance) is used as an estimate of the contrast of each image. Third, the image is normalized by its own standard deviation as an estimate of the structure comparison. Finally, the three components are combined to produce an overall similarity measure. Therefore, the structural similarity between image signal $\textbf{\emph{X}}$ and image signal $\textbf{\emph{Y}}$ can be modeled as
\begin{equation}
{\rm SSIM}(\textbf{\emph{X}},\textbf{\emph{Y}})=\frac{1}{m}\sum_{n=1}^m[L(X_n,Y_n)^{\alpha}C(X_n,Y_n)^{\beta}S(X_n,Y_n)^{\gamma}],
\end{equation}
where $m$ is the number of pixels; $L$, $C$, and $S$ are the luminance, contrast, and structure of the image, respectively; hyper-parameters $\alpha$, $\beta$ and $\gamma$ are used to weight the relative importance of $L$, $C$, and $S$, respectively; the default setting is $\alpha$ = $\beta$ = $\gamma$ = 1.

Based on SSIM measurement system, perceptual adversarial similarity score (PASS) is proposed to quantify human perception of AEs \cite{Rozsa2016}. The PASS between $\textbf{\emph{X}}$ and $\textbf{\emph{X}}^{'}$ is defined as
\begin{equation}
{\rm PASS}(\textbf{\emph{X}}^{'},\textbf{\emph{X}})={\rm SSIM}(\psi(\textbf{\emph{X}}^{'},\textbf{\emph{X}}),\textbf{\emph{X}}),
\end{equation}
where $\psi(\textbf{\emph{X}}^{'},\textbf{\emph{X}})$ represents the homography transform (a mapping from one plane to another) from the original image $\textbf{\emph{X}}$ to the adversarial image $\textbf{\emph{X}}^{'}$. PASS can quantify the AEs by measuring the similarity of the original image and the adversarial image. An appropriate PASS threshold can be set to distinguish the AEs with excessive perturbations. Meanwhile, attackers can also use the PASS threshold to optimize the methods of constructing AEs. Therefore, constructing an AE should satisfy
\begin{equation}
\mathop{\arg \min}_{d(\textbf{\emph{X}},\textbf{\emph{X}}^{'})} \ \textbf{\emph{X}}^{'}:f(\textbf{\emph{X}}^{'}) \neq y \ {\rm and} \ {\rm PASS}(\textbf{\emph{X}},\textbf{\emph{X}}^{'})\geq \theta,
\end{equation}
where $d(\textbf{\emph{X}},\textbf{\emph{X}}^{'})$ is some dissimilarity measure, $f(\textbf{\emph{X}}^{'})\neq y$ represents that the AE is misclassified by the model, $\theta$ is the PASS threshold set by the attacker, ${\rm PASS}(\textbf{\emph{X}},\textbf{\emph{X}}^{'})\geq \theta$ represents that the AE is not recognized by the human eyes.

\subsection{Adversarial Abilities and Adversarial Goals}

Adversarial ability is determined by how well attackers understand the model. Threat models in deep learning system are classified into the following types according to the attacker's abilities.

\textbf{White-box attack}. Attackers know everything related to trained neural network models, including training data, model parameters and model architectures.


\textbf{Grey-box attack}. Attackers know some model information such as model architectures, learning rate, training data and training steps, except model parameters. This attack is a byproduct of black-box attack and is not common in practical applications.

\textbf{Black-box attack}. Attackers do not know the architecture and parameters of the ML model, but can interact with the ML system. For example, the outputs can be determined by classifying random test vectors. Attackers utilize the transferability of AE to train an alternative model to construct AEs first, and then use the generated AEs to attack the unknown target model.

\begin{figure}
\centerline{{\includegraphics[width=\linewidth]{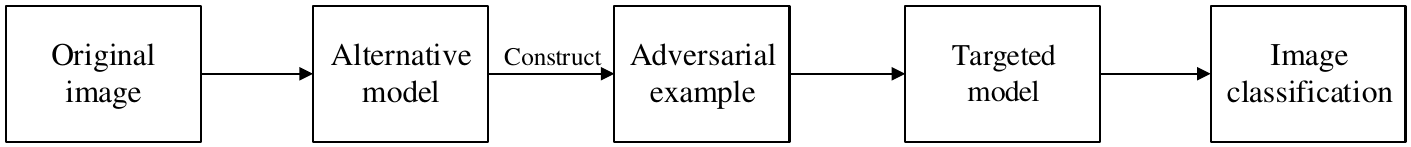}}}
\caption{\textbf{Black-box attack.} Alternative model: attackers know the model architecture and parameters; targeted model: attackers do not know the model architecture and parameters.}
\label{Fig8}
\end{figure}

The process of black-box attack is shown in Fig. \ref{Fig8}. First, attackers use the known dataset to train an alternative model. Then, attackers construct the corresponding AEs through the alternative model. Finally, these AEs can be used to attack the unknown target model due to the transferability. However, in some scenarios such as machine learning as a service, it is difficult to obtain the structure and parameters of the target model and training dataset. Papernot \emph{et al.} \cite{Papernot2016a} proposed a practical black-box attack method to generate AEs. The specific process is as follows:
\begin{enumerate}[  1)]
  \item Attackers use the target model as an oracle to construct a synthetic dataset, where the inputs are synthetically generated and the outputs are labels observed from the oracle.
\end{enumerate}
\begin{enumerate}[  2)]
  \item Attackers select a ML algorithm randomly and use the synthetic dataset to train a substitute model.
\end{enumerate}
\begin{enumerate}[  3)]
  \item Attackers generate AEs through the substitute model.
\end{enumerate}
\begin{enumerate}[  4)]
  \item Based on the transferability, attackers use the generated AEs to attack the unknown target model.
\end{enumerate}

The goal of adversarial deep learning is to misclassify the model. According to the different influence of the perturbation on the classifier, we classify the adversarial goals into four types:
\begin{enumerate}[  (1)]
  \item \textbf{Confidence reduction:} reduce the confidence of output classification.
\end{enumerate}

\begin{enumerate}[  (2)]

  \item \textbf{Non-targeted misclassification:} alter the output classification to any class which is different from the original class.
\end{enumerate}

\begin{enumerate}[  (3)]
  \item \textbf{Targeted misclassification:} force the output classification to be the specific target class.
\end{enumerate}

\begin{enumerate}[  (4)]
  \item \textbf{Source/Target misclassification:} select a specific input to generate a specific target class.
\end{enumerate}

\begin{figure}
\centerline{{\includegraphics[width=\linewidth]{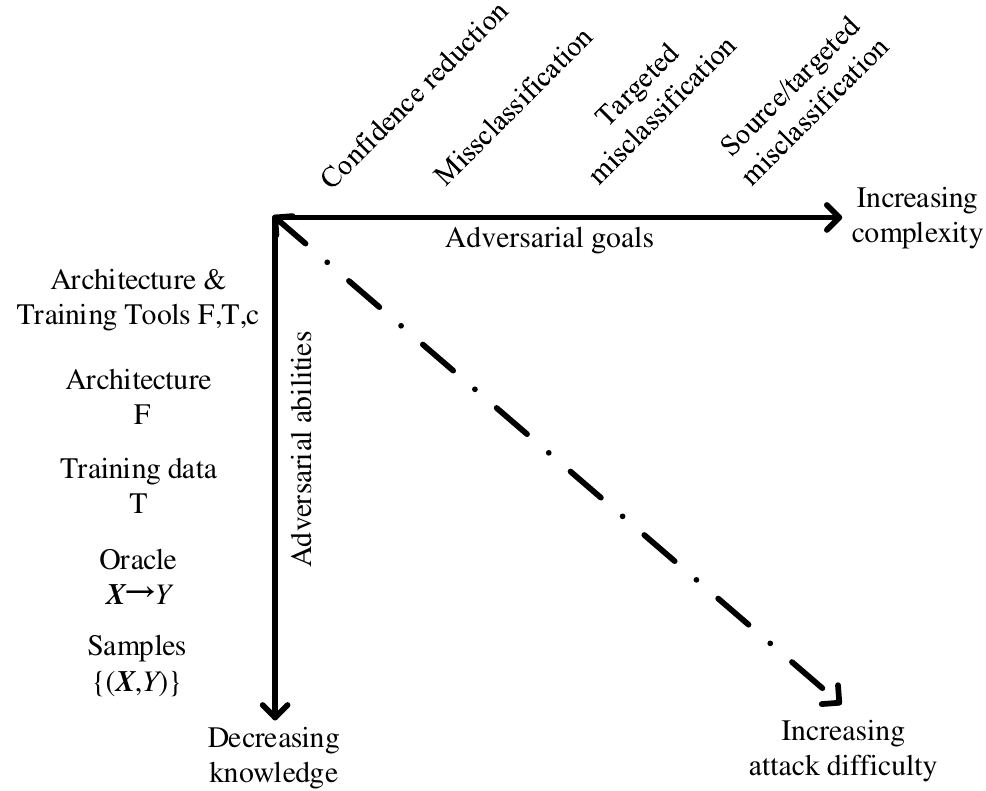}}}
\caption{Adversarial abilities and adversarial goals \cite{Papernot2016b}.}
\label{Fig9}
\end{figure}

As shown in Fig. \ref{Fig9}, the vertical axis represents the adversarial abilities which include architecture, training data, oracle (the adversary can obtain output classifications from provided inputs and observe the relationship between changes in inputs and outputs to adaptively construct AEs), and samples (the adversary has the ability to collect input and output pairs, but cannot modify these inputs to observe the difference in the output). The horizontal axis represents the adversarial goals, and the increasing complexity from left to right is confidence reduction, non-targeted misclassification, targeted misclassification, and source/target misclassification. In general, the weaker the adversarial ability or the higher the adversarial goal, the more difficult it is for the model to be attacked.

\section{Methods of Constructing AEs}
In what follows, several typical AE construction methods will be introduced in detail.

\subsection{Mainstream Attack Methods}

\begin{enumerate}
  \item [1)]{\emph{L-BFGS}}
\end{enumerate}

Szegedy \emph{et al.} \cite{Szegedy2013} proposed L-BFGS (Limited-memory; Broy-den, Fletcher, Goldforb, Shanno) to construct AEs. Given an image $\textbf{\emph{X}}$, attackers construct an image $\textbf{\emph{X}}^{'}$ similar to \textbf{\emph{X}} with $L_2$-norm and $\textbf{\emph{X}}^{'}$ can be labeled as a different class. The optimization problem is
\begin{equation}
\min \ ||\textbf{\emph{X}}-\textbf{\emph{X}}^{'}||_2,
\end{equation}
where $||\textbf{\emph{X}}-\textbf{\emph{X}}^{'}||_2$ is $L_2$-norm. The goal of the attack is to make $f(\textbf{\emph{X}}^{'})=\emph{l},\textbf{\emph{X}}^{'}\in [0,1]^n$, where $l$ is the target class. $f(\textbf{\emph{X}}^{'})=l$ is the nonlinear and non-convex function which is difficult to be solved directly. Therefore, the box-constrained L-BFGS is used for approximately solving the following problem:
\begin{equation}
\min\ c\cdot||\textbf{\emph{X}}-\textbf{\emph{X}}^{'}||_2+{\rm loss}_{F,l}(\textbf{\emph{X}}^{'}),
\end{equation}
where $c$ is a randomly initialized hyper-parameter, which is determined by linear search; ${\rm loss}_{F,l} (*)$ is the loss function. $f(\textbf{\emph{X}}^{'})=l$ is approximated by minimizing the loss function. Although this method is high stability and effectiveness, the calculation is complicated.

\begin{enumerate}
  \item [2)]{\emph{FGSM}}
\end{enumerate}

Goodfellow \emph{et al.} \cite{Goodfellow2014} proposed a simplest and fastest method to construct AEs, named fast gradient sign method (FGSM). The generated images are misclassified by adding perturbations and linearizing the cost function in the gradient direction. Given an original image \textbf{\emph{X}}, the problem can be solved with
\begin{equation}
\textbf{\emph{X}}^{\rm adv}=\textbf{\emph{X}}+\epsilon\, {\rm sign}(\nabla_X J(\textbf{\emph{X}},y_{\rm true})),
\end{equation}
where $\textbf{\emph{X}}^{\rm adv}$ represents an AE from $\textbf{\emph{X}}$, $\epsilon$ is a randomly initialized hyper-parameter, ${\rm sign}(*)$ is a sign function, $y_{\rm true}$ is the true label corresponding to $\textbf{\emph{X}}$, and $J(*)$ is the cost function used to train the neural network, $\nabla_X J(*)$ represents the gradient of $\textbf{\emph{X}}$.

There are two main differences between FGSM and L-BFGS. First, FGSM is optimized with the $L_{\infty}$-norm. Second, FGSM is a fast AE construction method because it does not require an iterative procedure to compute AEs. Hence it has lower computation cost than other methods. However, FGSM is prone to label leaking. Therefore, Kurakin \emph{et al.} \cite{Kurakin2016a} proposed FGSM-pred which uses the predicted label $y_{\rm pred}$ instead of true label $y_{\rm true}$. Researchers \cite{Kurakin2016a} also use the gradients with $L_2$- and $L_{\infty}$-norm, i.e., ${\rm sign}(\nabla_X J(\textbf{\emph{X}},y_{\rm true}))$ is changed to $({\nabla_X J(\textbf{\emph{X}},y_{\rm true})})/({||{\nabla_X J(\textbf{\emph{X}},y_{\rm true})}||_2})$ and $({\nabla_X J(\textbf{\emph{X}},y_{\rm true})})/({||{\nabla_X J(\textbf{\emph{X}},y_{\rm true})}||_{\infty}})$, these two methods are named as Fast grad.$L_2$ and Fast grad.$L_{\infty}$, respectively.

\begin{figure}
\centerline{{\includegraphics[width=3.5in,height=2.5in]{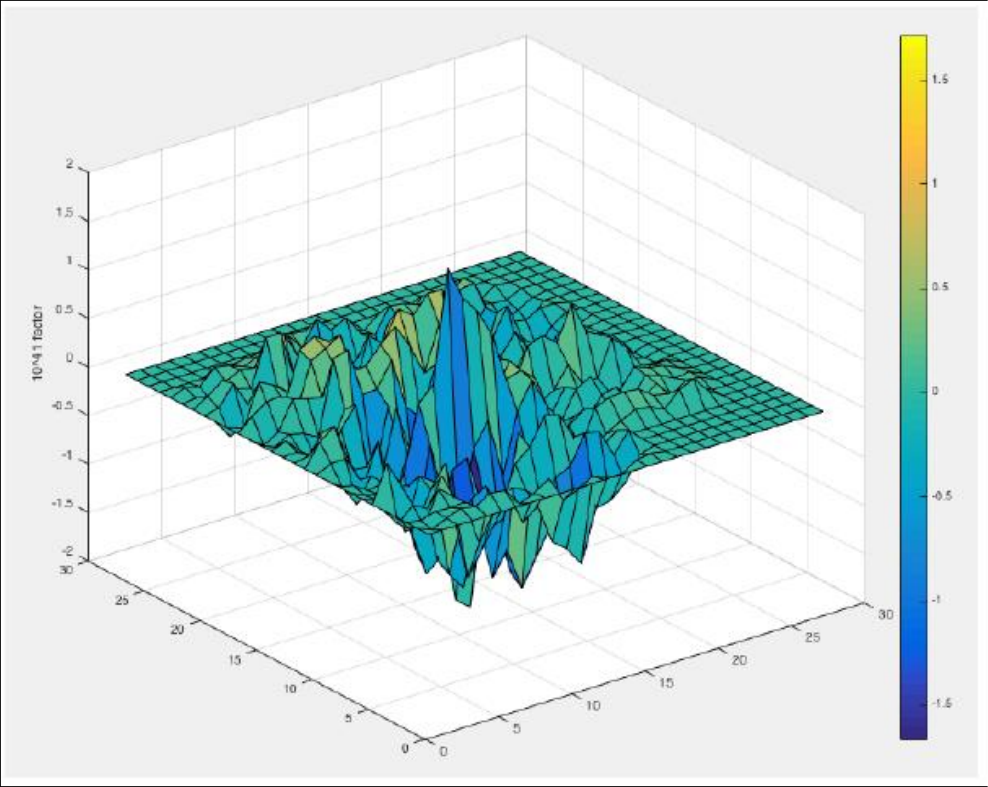}}}
\caption{Saliency map S with the 28$\times$28 image pixel, large absolute values correspond to features with a significant impact on the output when perturbed the input \cite{Papernot2016b}.}
\label{Fig10}
\end{figure}

\begin{enumerate}
  \item [3)]{\emph{IGSM}}
\end{enumerate}

It is difficult for FGSM to control the perturbation in constructing AEs. Kurakin \emph{et al.} \cite{Kurakin2016} proposed an optimized FGSM, named iterative gradient sign method (IGSM), which applies perturbations to multiple smaller steps and clips the results after each iteration to ensure that the perturbations are within the neighborhood of the original image. For the $\emph{N}$th iteration, the update process is
\begin{equation}
\begin{split}
\textbf{\emph{X}}_0^{\rm adv}=\textbf{\emph{X}},\ \textbf{\emph{X}}_{N+1}^{\rm adv}={\rm Clip}_{X,\epsilon}\{\textbf{\emph{X}}_N^{\rm adv}+ \\
\alpha\, {\rm sign}(\nabla_X J(\textbf{\emph{X}}_N^{\rm adv},y_{\rm true}))\},
\end{split}
\end{equation}
where ${\rm Clip}_{X,\epsilon}(*)$ denotes $[\textbf{\emph{X}}-\epsilon,\textbf{\emph{X}}+\epsilon]$.

IGSM is non-linear in the gradient direction and requires multiple iterations, which is simpler than L-BFGS method in calculation, and the success rate of AE construction is higher than FGSM. IGSM can be further divided into two types: 1) reducing the confidence of the original prediction as the original class; 2) increasing the confidence of the prediction that originally belongs to the class with the smallest probability.

Recently, Dong \emph{et al.} \cite{Dong} proposed a momentum iterative method (MIM). The basic idea is to add momentum based on the IGSM. The weakness of previous iterative attacks is that the transferability (black-box attack) is weakened when the number of iterations increases, which can be addressed after adding momentum. MIM attack not only enhances the attack ability on the white-box model, but also increases the success rate on the black-box model. The momentum iterative gradient sign method for the targeted attack is given by
\begin{equation}
\begin{array}{lr}
\begin{aligned}
& \textbf{\emph{g}}_{t+1}=\mu \textbf{\emph{g}}_t + \frac{\nabla_X J(\textbf{\emph{X}}_t^{\rm adv},y_{\rm target})}{||{\nabla_X J(\textbf{\emph{X}}_t^{\rm adv},y_{\rm target})}||_1},   \hfill\\
& \textbf{\emph{X}}_{t+1}^{\rm adv}={\rm Clip}_{X,\epsilon}(\textbf{\emph{X}}_t^{\rm adv} + \alpha\, {\rm sign}(\textbf{\emph{g}}_{t+1})),\hfill
\end{aligned}
\end{array}
\end{equation}
where $\textbf{\emph{g}}_t$ gathers the gradients of the first $t$ iterations with a decay factor $\mu$, and its initial value is 0.

\begin{enumerate}
  \item [4)]{$Iter l.l$}
\end{enumerate}

FGSM and L-BFGS try to increase the probability of predicting wrong results but do not specify which wrong class should be selected by the model. These methods are sufficient for small datasets such as Mixed National Institute of Standards and Technology Database (MNIST) and CIFAR-10. On ImageNet, with a larger number of classes and varying degrees of significance between classes, FGSM and L-BFGS may construct uninteresting misclassifications, such as misclassifying one type of cat into another cat. To generate more meaningful AEs, a novel AE generation method is proposed by perturbing the target class with the lowest probability so that this least-likely class turns to become the correct class after the perturbation, which is called as iterative least-likely class method ($iter l.l$) \cite{Kurakin2016}. To make the adversarial image $\textbf{\emph{X}}^{\rm adv}$ be classified as $y_{LL}$, we have the following procedure:
\begin{equation}
\begin{split}
\textbf{\emph{X}}_0^{\rm adv}=\textbf{\emph{X}},\ \textbf{\emph{X}}_{N+1}^{\rm adv}={\rm Clip}_{X,\epsilon}\{\textbf{\emph{X}}_N^{\rm adv}- \\
\alpha\, {\rm sign}(\nabla_X J(\textbf{\emph{X}}_N^{\rm adv},y_{LL}))\},
\end{split}
\end{equation}
where $y_{LL}$ represents the least likely (the lowest probability) target class. For a classifier with good performance, the least likely class is usually quite different from the correct class. Therefore, this attack method can lead to some interesting errors, such as misclassifying a cat as an aircraft. It is also possible to use a random class as the target class, which is called as iteration random class method.


\begin{enumerate}
  \item [5)]{\emph{JSMA}}
\end{enumerate}

Papernot \emph{et al.} \cite{Papernot2016b} proposed the Jacobian-based Saliency Map Attack (JSMA), which is based on the $L_0$ distance norm. The basic idea is to construct a saliency map with the gradients and model the gradients based on the impact of each pixel. The gradients are directly proportional to the probability that the image is classified as the target class, i.e., changing a pixel with a larger gradient will significantly increase the likelihood that the model classifies the image as the target class. JSMA allows us to select the most important pixel (the maximum gradient) based on the saliency map and then perturb the pixel to increase the likelihood of labeling the image as the target class. More specifically, JSMA includes the following steps:

\begin{enumerate}[  (1)]
  \item Compute forward derivative $\nabla F(\textbf{\emph{X}})$.
  \begin{equation}
  \nabla F(\textbf{\emph{X}})=\frac{\partial F(\textbf{\emph{X}})}{\partial\textbf{\emph{X}}}={[\frac{\partial F_j(\textbf{\emph{X}})}{\partial X_i}]}_{i\in1...M,j\in1...N}.
  \end{equation}
\end{enumerate}

\begin{enumerate}[  (2)]
  \item Construct a saliency map $S$ based on the forward derivative, as shown in Fig. \ref{Fig10}.
\end{enumerate}

\begin{enumerate}[  (3)]
  \item Modify the most important pixel based on the saliency map, repeat this process until the output is the target class or the maximum perturbation is got.
\end{enumerate}

When the model is sensitive to the change of inputs, JSMA is easier to calculate the minimum perturbation to generate the AEs. JSMA has high computational complexity while the generated AEs have a high success rate and transfer rate.

\begin{enumerate}
  \item [6)]{\emph{DeepFool}}
\end{enumerate}

Mohsen \emph{et al.} \cite{Moosavi-Dezfooli2016} proposed a non-targeted attack method based on the $L_2$-norm, called DeepFool. Assuming that the neural network is completely linear, there must be a hyperplane separating one class from another. Based on this assumption, we analyze the optimal solution to this problem and construct AEs. The corresponding optimization problem is
\begin{equation}
r_*(\textbf{\emph{X}}_0)=\arg \min {||\textbf{\emph{r}}||}_2,
\end{equation}
subject to ${\rm sign}(f(\textbf{\emph{X}}_0+\textbf{\emph{r}}))\neq {\rm sign}(f(\textbf{\emph{X}}_0))$, where \textbf{\emph{r}} indicates the perturbation.

\begin{table*}
\caption{Advantages and Disadvantages of Different Attack Methods}
\label{Table2}
\centering
\begin{tabular}{|c|c|c|}

\hline
\thead{Method} & \thead{Advantage} & \thead{Disadvantage}
 \\
\hline
\thead{L-BFGS \cite{Szegedy2013}} & \thead{High stability and effectiveness} & \thead{High computational and time complexity}   \\
\hline
\thead{FGSM \cite{Goodfellow2014} } & \thead{Low computational complexity, high transfer rate} & \thead{Low success rate, label leaking}  \\
\hline
\thead{IGSM \cite{Kurakin2016} } & \thead{Small perturbations, high success rate} & \thead{Low transfer rate, low success rate for balck-box attacks} \\
\hline
\thead{Iter.l.l \cite{Kurakin2016}} & \thead{Small perturbations, high success rate} & \thead{Low transfer rate, low success rate for balck-box attacks} \\
\hline
\thead{JSMA \cite{Papernot2016b}} & \thead{Small perturbations, high success rate} & \thead{High computational complexity}\\
\hline
\thead{Uni.perturbations \cite{Moosavi-Dezfooli2017}} & \thead{High generalization ability, high transfer rate} & \thead{The perturbation is not easy to control,\\ low success rate for target attacks} \\
\hline
\thead{DeepFool \cite{Moosavi-Dezfooli2016} } & \thead{Low computational complexity, small perturbations} & \thead{Low success rate for balck-box attacks}\\
\hline
\thead{One-pixel \cite{Su2017} } & \thead{Low computational complexity} & \thead{Low success rate for target attacks, large perturbations} \\
\hline
\thead{CW Attack \cite{Carlini2017a}} & \thead{Small perturbations, high transfer rate and success rate} & \thead{High computational complexity}\\
\hline
\thead{Ensemble Attack \cite{Liu2016}} & \thead{Simple computation, good generalization} & \thead{Low success rate for balck-box attacks}\\
\hline
\end{tabular}
\end{table*}

\begin{figure}
\centerline{{\includegraphics[width=2.625in,height=1.5in]{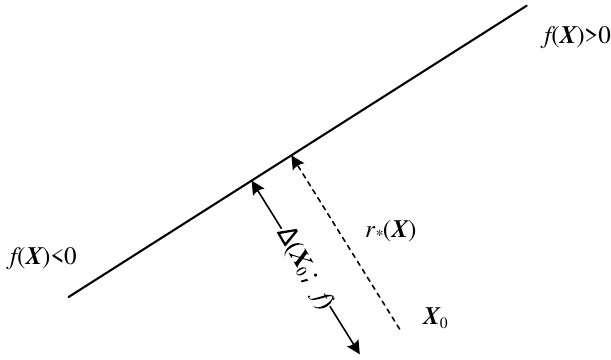}}}
\caption{AEs for a linear binary classifier \cite{Moosavi-Dezfooli2016}.}
\label{Fig11}
\end{figure}

As shown in Fig. \ref{Fig11}, $\textbf{\emph{X}}_0$ is the original example, $f(\textbf{\emph{X}})$ is a linear binary classifier, the straight line $\textbf{\emph{W}}\textbf{\emph{X}}+b=0$ is the decision boundary, and $r_*(\textbf{\emph{X}})$ is the distance from the original example to the decision boundary, i.e., the distance from $\textbf{\emph{X}}_0$ to the straight line $\textbf{\emph{W}}\textbf{\emph{X}}+b=0$. The distance is equivalent to the perturbation $\Delta(\textbf{\emph{X}}_0; f)$. Therefore, when $\Delta(\textbf{\emph{X}}_0; f)> r_*(\textbf{\emph{X}})$, the AE can be generated.

Compared with L-BFGS, DeepFool is more efficient and powerful. The basic idea is to find the decision boundary that is the closest to $\textbf{\emph{X}}$ in the image space, and then use the boundary to fool the classifier. It is difficult to solve this problem directly in neural networks with high dimension and nonlinear space. Therefore, a linearized approximation is used to iteratively solve this problem. The approximation is to linearize the intermediate $\textbf{\emph{X}}_0$ classifier  in each iteration, and obtain an optimal update direction on the linearized model. Then $\textbf{\emph{X}}_0$ is iteratively updated in this direction by a small step $\alpha$, repeating the linear update process until $\textbf{\emph{X}}_0$ crosses the decision boundary. Finally, AEs can be constructed with subtle perturbations.
\begin{enumerate}
  \item [7)]{\emph{Universal Adversarial Perturbations}}
\end{enumerate}

FGSM, JSMA and DeepFool can only generate adversarial perturbations to fool a network on a single image. Moosavi-Dezfooli \emph{et al.} \cite{Moosavi-Dezfooli2017} proposed a universal image-agnostic perturbation attack method which fools classifiers by single adversarial perturbation to all images. The specific problem can be defined as finding a universal perturbation $\textbf{\emph{v}}$, such that $\hat{k}(\textbf{\emph{X}}+\textbf{\emph{v}})\neq \hat{k}(\textbf{\emph{X}})$ for most examples in the dataset subject to the distribution $\mu$ and can be expressed as
\begin{equation}
\hat{k}(\textbf{\emph{X}}+\textbf{\emph{v}})\neq \hat{k}(\textbf{\emph{X}})\quad {\rm for}\ {\rm ``most"}\ \textbf{\emph{X}} \sim \mu,
\end{equation}
where $\hat{k}$ denotes a classification function that labels each image. Such perturbation $\textbf{\emph{v}}$ is called as the universal perturbation. Attackers' goal is to find the $\textbf{\emph{v}}$ that satisfies the following two constraints:

\begin{equation}
\left\{
\begin{array}{lr}
\begin{aligned}
&1.\;||\textbf{\emph{v}}||_p \leq \S, \hfill\\
&2.\mathop{\mathbb{P}}_{\textbf{\emph{X}}\sim\mu}\left(\;\hat{k}(\textbf{\emph{X}}+\textbf{\emph{v}})\neq\hat{k}(\textbf{\emph{X}})\right)\geq
{1-\delta},\hfill
\end{aligned}
\end{array}
\right.
\end{equation}
where $||\cdot||_p $  denotes the $p$-norm, the parameter $\S$  controls the magnitude of the perturbation $\textbf{\emph{v}}$, and $\delta$  quantifies the desired fooling rate for all images. The attack method has two characteristics: 1) the perturbation is related to the target model rather than the image; 2) the small perturbation will not change the structure of the image itself.

\begin{enumerate}
  \item [8)]{\emph{Carlini and Wagner (CW) Attack (C$\&$W)}}
\end{enumerate}

Carlini and Wagner \cite{Carlini2017a} proposed a powerful attack method based on L-BFGS. The attack with $L_0, L_2, L_{\infty}$ distance norm can be targeted or non-targeted, and we take the non-targeted $L_2$-norm as an example here. The corresponding optimization problem is
\begin{equation}
\min\ ||\boldsymbol{\delta}||_2+c\cdot f(\textbf{\emph{X}}+\boldsymbol{\delta}),
\end{equation}
where $\textbf{\emph{X}}+\boldsymbol{\delta} \in[0,1]^n$, $c$ is a hyper-parameter that can balance these two terms, and $\boldsymbol{\delta}$ is a small perturbation. The objective function $f(\textbf{\emph{X}}^{'})$ is defined as
\begin{equation}
f(\textbf{\emph{X}}^{'})=\max(\max\{{{Z(\textbf{\emph{X}}^{'})}_i}:i\neq t\}-Z({\textbf{\emph{X}}^{'})}_t,-l),
\end{equation}
where $Z(\textbf{\emph{X}}^{'})$ is the last hidden layer, $t$ is the true label, and $l$ is a hyper-parameter, which is used to control the confidence level of the model misclassification, and the AE $\textbf{\emph{X}}^{'}$ can be classified as $t$ with high confidence by adjusting the value of $l$. In general, high confidence attacks have large perturbations and high transfer rates, and $CW$ Attack based on the $L_0, L_2, L_{\infty}$ distance metric can defeat the defensive distillation \cite{Papernot2016}. There are three improvements to this attack based on L-BFGS:
\begin{itemize}
  \item Use the gradient of the actual output in the model instead of the gradient of softmax.
  \item Apply different distance metrics ($L_0, L_2, L_{\infty}$).
  \item Apply different objective functions $f(\textbf{\emph{X}}^{'})$.
\end{itemize}

\begin{enumerate}
  \item [9)]{\emph{Ensemble Attack}}
\end{enumerate}

Liu \emph{et al.} \cite{Liu2016} proposed an ensemble attack method combining multiple models to construct AEs. If an adversarial image remains adversarial for multiple models, it is likely to be transferred to other models. Formally, given $k$ white-box models with softmax outputs being $J_1,...,J_k$, an original image $\textbf{\emph{X}}$ and its true label $y$, the ensemble-based approach solves the following optimization problem (for targeted attack):
\begin{equation}
\mathop{\arg \min}_{\textbf{\emph{X}}^{'}}-\log((\sum_{i=1}^k \alpha_i J_i(\textbf{\emph{X}}^{'}))\cdot \textbf{1}_{y^{'}})+\lambda d(\textbf{\emph{X}},\textbf{\emph{X}}^{'}),
\end{equation}
where $y^{'}$ is the target label specified by the attacker, $\sum_{i=1}^k \alpha_i J_i(\textbf{\emph{X}}^{'})$ is the ensemble model, and $\alpha_i$ is the weight of the $i$-th model, $\sum_{i=1}^k \alpha_i=1$, $\lambda$ is a randomly initialized parameter that is used to control the weight of the two terms. The goal is to ensure the generated AEs are still adversarial for the other black-box model $J_{k+1}$. Since the decision boundaries for different models are almost the same, the transferability of targeted AEs is improved significantly.

\subsection{Other Attack Methods}
From the perspective of attackers, the goal of the attack is to construct strong AEs with small perturbations and fool the model with high confidence without being recognized by human eyes. Recently, in addition to typical AE construction methods introduced above, a lot of other attack methods have been proposed \cite{Yuan2019,Liu2018}.
Xia \emph{et al.} \cite{Xia2016} proposed AdvGAN to construct AEs. The basic idea is to use generative adversarial networks to construct targeted AEs, which not only learns and preserves the distribution of the original examples, but also guarantees the diversity of perturbations and enhances generalization ability significantly.
Tramèr \emph{et al.} \cite{Tramer} proposed an ensemble attack RAND + FGSM. First, they added a small random perturbation RAND to ``escape'' the non-smooth vicinity of the data point before computing the gradients. Then, they applied the FGSM to enhance the attack ability greatly. Compared with the FGSM, this method has a higher success rate and can effectively avoid label leaking.
Su \emph{et al.} \cite{Su2017} proposed a one pixel attack method which only changes one pixel for each image to construct AEs to fool DNNs. However, such simple perturbation can be recognized by human eyes easily.
Weng \emph{et al.} \cite{Weng2018} proposed a computationally feasible method called Cross Lipschitz Extreme Value for nEtwork Robustness (CLEVER), which applies extreme value theory to estimate a lower bound of the minimum adversarial perturbation required to misclassify the image. CLEVER is the first attack-independent method and can evaluate the intrinsic robustness of neural networks.
Brown \emph{et al.} \cite{Brown2017} proposed adversarial patch which does not need to subtly transform an existing image into another and can be placed anywhere within the field of view of the classifier to cause the classifier to output a targeted class. Li \emph{et al.} \cite{Li2019} studied the security of real-world cloud-based image detectors, including Amazon Web Services (AWS), Azure, Google Cloud, Baidu Cloud and Alibaba Cloud. Specifically, they proposed four different attacks based on semantic segmentation, which generate semantics-aware AEs by interacting only with the black-box application programming interfaces (APIs).

As discussed above, many AE attacks have been proposed in recent years. We summarized the advantages and disadvantages of several typical attack methods in Table \ref{Table2}.

\section{Comparison of Various Attack Methods}
In this section, we compare the attributes of different attack methods in terms of black/white-box, attack type, targeted/non-targeted and PASS. Then we conduct a lot of experiments to compare the success rate and the transfer rate for different attack methods. The code to reproduce our experiments is available online at \textcolor[rgb]{0.00,0.07,1.00}{http://hardwaresecurity.cn/SurveyAEcode.zip.}

\begin{table*}
\caption{Comparison of Different Attack Methods on Attributes}
\label{Table1}
\centering
\begin{tabular}{|c|c|c|c|c|}

\hline
\thead{Method} & \thead{Black/White-box} & \thead{Attack Type} & \thead{Targeted/Non-targeted} & \thead{PASS}
 \\
\hline
\thead{L-BFGS \cite{Szegedy2013}} & \thead{White-box} & \thead{Gradient-based} & \thead{Targeted} & \thead{$\ast$}  \\
\hline
\thead{FGSM \cite{Goodfellow2014} } & \thead{White-box} & \thead{Gradient-based} & \thead{Non-targeted} & \thead{$\ast\ast\ast\ast\ast$} \\
\hline
\thead{IGSM \cite{Kurakin2016} } & \thead{White-box} & \thead{Gradient-based}& \thead{Targeted/Non-targeted} & \thead{$\ast\ast$} \\
\hline
\thead{Iter.l.l \cite{Kurakin2016}} & \thead{White-box} & \thead{Gradient-based}& \thead{Non-targeted} & \thead{$\ast\ast$} \\
\hline
\thead{JSMA \cite{Papernot2016b}} & \thead{White-box} & \thead{Gradient-based}& \thead{Targeted} & \thead{$\ast\ast$}\\
\hline
\thead{Uni.perturbations \cite{Moosavi-Dezfooli2017}} & \thead{White-box} & \thead{Decision boundary-based} & \thead{Non-targeted} & \thead{$\ast\ast\ast$}\\
\hline
\thead{DeepFool \cite{Moosavi-Dezfooli2016} } & \thead{White-box} & \thead{Decision boundary-based}& \thead{Non-targeted} & \thead{$\ast$}\\
\hline
\thead{One-pixel \cite{Su2017} } & \thead{Black-box} & \thead{-}& \thead{Non-targeted} & \thead{$\ast\ast\ast\ast$}\\
\hline
\thead{CW Attack \cite{Carlini2017a}} & \thead{White-box} & \thead{Iterative optimization}& \thead{Targeted/Non-targeted} & \thead{$\ast$}\\
\hline
\thead{Ensemble Attack \cite{Liu2016}} & \thead{White-box} & \thead{Ensemble optimization}& \thead{Non-targeted} & \thead{$\ast\ast\ast$}\\
\hline
\end{tabular}
\end{table*}

\subsection{Experimental Setup}
\textbf{Platform}. All the experiments are conducted on a machine equipped with an AMD Threadripper 1920X CPU, a NVIDIA GTX 1050Ti GPU and 16G memory, and implemented in tensorflow-gpu 1.6.0 with Python 3.6.

\textbf{Dataset}. All of the experiments described in this section are performed on ILSVRC 2012 dataset \cite{Russakovsky2015}. The ILSVRC 2012 dataset is a subset of ImageNet, containing about 1000 images of 1000 categories. The examples are split among a training set of approximately 1.2 million examples, a validation set of 50,000 and a test set of 150,000. In our experiments, we randomly select 1000 images from 1000 categories in the ILSVRC 2012 validation set. The size of each image is $299 \times 299 \times 3$. The intensity values of pixels in all these images are scaled to a real number in [0, 1].

\textbf{DNN Model}. For the dataset, we evaluate the success rate and transfer rate of AEs on five popular deep network architectures: Inception V3 \cite{Szegedy2015}, AlexNet \cite{Krizhevsky2017}, ResNet34 \cite{He2016}, DenseNet20 \cite{Huang2017a} and VGG19 \cite{Simonyan2014}, which are pretrained by the ImageNet dataset.

\subsection{Comparison of Different Attack Methods on Attributes}

Table \ref{Table1} summarizes the attribute information of mainstream attacks. We find that: 1) most of the attack methods are white-box attacks. In the scenario of black-box attacks, attackers are difficult to construct AE attacks on ML models; 2) most of the mainstream attacks are based on the gradient. Later on, other attacks such as decision boundary, iterative optimization and ensemble optimization are proposed; 3) the more asterisks, the larger the value of PASS, i.e., the generated AEs are easier to be perceived by human eyes. For example, since FGSM is prone to the label leakage effect, the best perturbation to generate AEs is large (i.e., the PASS is large) and hence the generated AEs are easier to be perceived.

\subsection{The Success Rate}
The success rate of various AE construction methods for targeted and non-targeted attacks on the Inception V3 model is evaluated. As shown in Table \ref{Table4}, the success rate of targeted attacks is lower than the non-targeted attacks for the same AE construction method, and MIM has the highest success rate for target and non-target attacks. Note that JSMA needs to calculate the forward derivative of each pixel in the image to construct the Jacobian matrix. Therefore, JSMA is computationally expensive. In the experiment, we do not report the success rate and transfer rate for JSMA because it runs out of memory on the big dataset.

\begin{table}
\caption{The Success Rate of Adversarial Examples}
\label{Table4}
\centering
 \begin{threeparttable}
\begin{tabular}{|c|c|c|}
\hline
\diagbox{Method}{Type} & \thead{Targeted} & \thead{Non-targeted} \\
\hline
 \thead{L-BFGS \cite{Szegedy2013}} & \thead{96.1\%} & \thead{NA}  \\
\hline
 \thead{FGSM \cite{Goodfellow2014}} & \thead{NA} & \thead{74.6\%}  \\
\hline
 \thead{IGSM \cite{Kurakin2016}} & \thead{85.4\%} & \thead{99.1\%}  \\
\hline
 \thead{MIM \cite{Dong}} & \thead{99.3\%} & \thead{100\%}  \\
\hline
 \thead{DeepFool \cite{Moosavi-Dezfooli2016}} & \thead{NA} & \thead{100\%} \\

\hline
 \thead{CW Attack \cite{Carlini2017a}} & \thead{93.2\%} & \thead{98.4\%}  \\
\hline
 \thead{Uni.perturbations \cite{Moosavi-Dezfooli2017}} & \thead{NA} & \thead{88.2\%}  \\
 \hline
 \thead{JSMA$^*$\cite{Papernot2016b}} & \thead{-} & \thead{-}  \\
\hline
\end{tabular}
\begin{tablenotes}
 \footnotesize
\item[$*$] Out of Memory.
\end{tablenotes}
 \end{threeparttable}
\end{table}

\begin{table*}
\caption{The Transfer Rate of Adversarial Examples}
\label{Table3}
\centering
\begin{threeparttable}
\begin{tabular}{|c|c|c|c|c|c|c|}

\hline
\diagbox{Method}{Model} & \thead{Targeted/Non-targeted} & \thead{A$\rightarrow$B} & \thead{A$\rightarrow$C} & \thead{A$\rightarrow$D} & \thead{A$\rightarrow$E}& \thead{Perturbations($P$)} \\
\hline
\thead{L-BFGS \cite{Szegedy2013}} & \thead{Targeted} & \thead{0.3\%} & \thead{0.3\%} & \thead{0.3\%} & \thead{0.2\%}& \thead{1.02} \\
\hline
\thead{FGSM \cite{Goodfellow2014} } & \thead{Non-targeted} & \thead{66.8\%} & \thead{51.9\%} & \thead{51.5\%} & \thead{64.8\%} & \thead{15.23}\\
\hline
\multirow{2}{*}{IGSM \cite{Kurakin2016}} & \thead{Targeted} & \thead{0.3\%}& \thead{0.2\%} & \thead{0.1\%}& \thead{0.1\%}& \thead{8.87}\\
  \cline{2-7}
   & \thead{Non-targeted} & \thead{39.9\%}& \thead{34.4\%} & \thead{33.7\%}& \thead{46.7\%} & \thead{8.85}\\
\hline
\multirow{2}{*}{MIM \cite{Dong}} & \thead{Targeted} & \thead{0\%}& \thead{0.4\%} & \thead{0.3\%}& \thead{0.5\%}& \thead{13.55}\\
  \cline{2-7}
   & \thead{Non-targeted} & \thead{56.4\%}& \thead{45.4\% } & \thead{44.8\%}& \thead{59.9\%}& \thead{13.54}\\
\hline

\thead{DeepFool \cite{Moosavi-Dezfooli2016} } & \thead{Non-Targeted} & \thead{30.5\%}& \thead{16.6\%} & \thead{10.5\%}& \thead{16.7\%}  & \thead{0.49}\\

\hline
\multirow{2}{*}{CW Attack \cite{Carlini2017a}}& \thead{Targeted} & \thead{0\%}& \thead{0\%} & \thead{0\%}& \thead{0\%} & \thead{0.78}\\
  \cline{2-7}
   & \thead{Non-targeted } & \thead{4.3\%}& \thead{3.9\%} & \thead{2.9\%}& \thead{5.3\% }& \thead{0.76}\\
\hline
\thead{Uni.perturbations \cite{Moosavi-Dezfooli2017}} & \thead{Non-targeted} & \thead{44.2\%}& \thead{29.1\%} & \thead{23.2\%}& \thead{35.4\%}  & \thead{6.01}\\
\hline
\thead{JSMA$^*$\cite{Papernot2016b}} & \thead{Targeted} & \thead{-}& \thead{-} & \thead{-}& \thead{-}  & \thead{-}\\
\hline
\end{tabular}
\begin{tablenotes}
 \footnotesize
\item[$*$] Out of Memory.
\end{tablenotes}
 \end{threeparttable}
\end{table*}

\subsection{The Transfer Rate}
The transfer rates of typical AEs for targeted and non-targeted attacks among the five models (Inception V3 (A), AlexNet (B), ResNet (C), DenseNet (D) and VGG (E)) are shown in Table \ref{Table3}. Inception V3 is used as the source model and the other models are used as the target models. We collect 1000 adversarial images generated by each AE construction method in Inception V3 and apply them to other models to test the transfer rate of AEs. For each AE construction method, we define the average perturbation $P$ as
\begin{equation}
P = \frac{\sum_{i=1}^N \sum_{r=1}^R \sum_{c=1}^C \sum_{l=1}^L {\rm abs}({\rm pixel}_{i,r,c,l}^{\rm ori} - {\rm pixel}_{i,r,c,l}^{\rm adv})}{N \times R \times C \times L},
\end{equation}
where $N$, $R$, $C$ and $L$ are the number of images, rows, columns and layers respectively, ${\rm pixel}_{i,r,c,l}^{\rm ori(adv)}$ is the pixel value of row $r$, column $c$, and layer $l$ of original (adversarial) image $i$. ${\rm abs}({\rm pixel}_{i,r,c,l}^{\rm ori} - {\rm pixel}_{i,r,c,l}^{\rm adv})$ is the absolute value of $({\rm pixel}_{i,r,c,l}^{\rm ori} - {\rm pixel}_{i,r,c,l}^{\rm adv})$. The magnitude of the perturbation is proportional to the transfer rate, i.e., the greater the perturbation, the higher the transfer rate. In addition, we find that the transfer rate of AE construction methods is low, i.e., the transfer rates of most of targeted attacks are less than 0.5$\%$, which means that it is difficult to conduct AE attacks in the real black-box scenario and it is urgent to develop AE attack methods with a high transfer rate.



\section{Defenses against Adversarial Examples}
AEs bring a great threat to the security-critical AI applications such as face payment \cite{Sharif}, medical systems \cite{Finlayson2019a} and autonomous vehicles \cite{Huang2018,Huang2017} based on image recognition in deep learning. Vulnerability to AEs is not unique to deep learning. All ML models are vulnerable to AEs \cite{Papernot2016a}. Therefore, defending against AEs is urgent for ML security. In this section, we will briefly describe the basic goals of defending against AEs, then detail the current defense techniques and their limitations. Finally, some suggestions are presented for future research work on the problems of the current defense techniques.

\subsection{Defense Goals}
Generally, there are four defense goals:
\begin{enumerate}[ \  1)]
\item \textbf{Low impact on the model architecture:} when constructing any defense against AEs, the primary consideration is the minimal modification to model architectures.
\end{enumerate}

 \begin{enumerate}[ \ 2)]
\item \textbf{Maintain model speed:} running time is very important for the availability of DNNs. It should not be affected during testing. With the deployment of defenses, DNNs should still maintain high performance on large datasets.
\end{enumerate}

\begin{enumerate}[  \ 3)]
\item \textbf{Maintain accuracy:} defenses should have little impact on the classification accuracy of models.
\end{enumerate}

\begin{enumerate}[  \ 4)]
\item \textbf{Defenses should be targeted:} defenses should be effective for the examples that are relatively close to the training set. Since the examples that are far from the dataset are relatively secure, the perturbations to these examples are easily detected by the classifier.
\end{enumerate}

\subsection{Current Defenses}

\begin{enumerate}
  \item [1)]{\emph{Adversarial Training}}
\end{enumerate}

AEs have been used to improve the anti-interference ability for AI models. In 2015, Goodfellow \emph{et al.} \cite{Goodfellow2014} proposed the adversarial training to improve the robustness of the model. The basic idea is to add AEs to the training data and continuously generate new AEs at each step of the training. The number and relative weight of AEs in each batch is controlled by the loss function independently. The corresponding loss function \cite{Kurakin2016a} is
\begin{equation}
\begin{split}
{\rm LOSS}=\frac{1}{(m-k)+\lambda k}(\sum_{i\in {\rm CLEAN}}L(X_i|y_i)+ \\
  \lambda\sum_{i\in {\rm ADV}}L(X_i^{\rm adv}|y_i)),
\end{split}
\end{equation}
where $L(\textbf{\emph{X}}|y)$ is a loss function of the example $\textbf{\emph{X}}$ with a true label $y$, $m$ is the total number of training examples, $k$ is the number of AEs, and $\lambda$ is a hyper-parameter used to control the relative weight of the AEs in the loss function. When $k = m/2$, i.e., when the number of AEs is the same as the number of original examples, the model has the best effect in the adversarial training.

Adversarial training is not the same as data augmentation. The augmented data may appear in the test set, while AEs are usually not shown in the test set but can reveal the defects of the model. Adversarial training can be viewed as the process of minimizing classification error rates when the data is maliciously perturbed. In the following two situations, it  is suggested to use adversarial training:
\begin{enumerate}
  \item [i)]\textbf{Overfitting:} when a model is overfitting, a regularization term is needed.
\end{enumerate}
\begin{enumerate}
  \item [ii)]\textbf{Security:} when AEs refer to security problems, adversarial training is the most secure method among all known defenses with only a small loss of accuracy.
\end{enumerate}

Although a model is robust to white-box attacks after adversarial training, it is still vulnerable to the AEs generated from other models, i.e., the model is not robust to black-box attacks. Based on this attribute, Tramèr \emph{et al.} \cite{Tramer} proposed the concept of ensemble adversarial training. The main idea is to augment the training data which constructed not only from the model being trained but also from the other pre-trained models, which increases the diversity of AEs and improves the generalization ability.

\begin{figure}
\centerline{{\includegraphics[width=\linewidth]{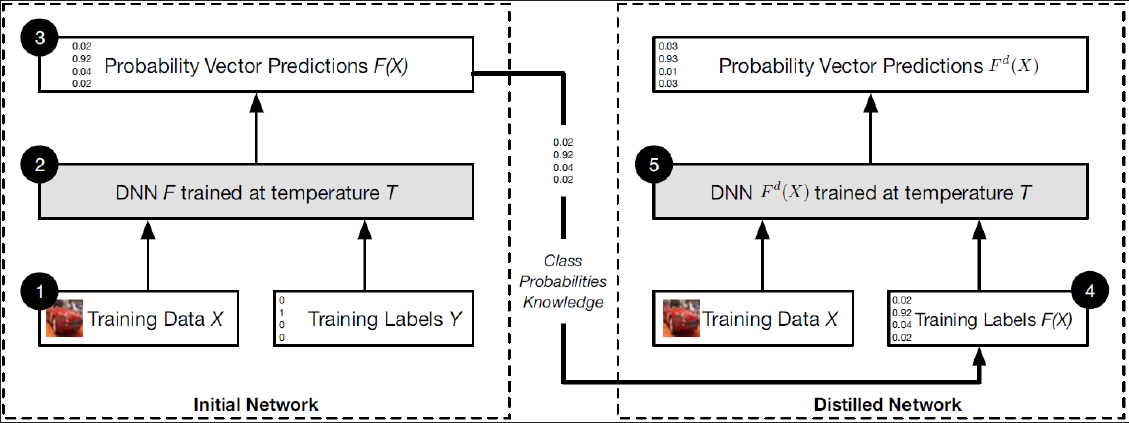}}}
\caption{\textbf{The pipeline of defensive distillation.} The initial network is trained at temperature $T$ on the training set ({\textbf{\emph{X}},$Y$(\textbf{\emph{X}})}), the distilled network is trained at the same temperature $T$ on the new training set ({\textbf{\emph{X}},$F$(\textbf{\emph{X}})}) \cite{Papernot2016}. }
\label{Fig13}
\end{figure}

\begin{enumerate}
  \item [2)]{\emph{Defensive Distillation}}
\end{enumerate}

Adversarial training needs AEs to train the model, thus the defense is related to the process of AEs construction. For any defense, the defense effect is quite different for different attack methods. In 2016, Papernot \emph{et al.} \cite{Papernot2016} proposed a universal defensive method for neural networks, which is called defensive distillation. The distillation method uses a small model to simulate a large and computationally intensive model without affecting the accuracy and can solve the problem of information missing. Different from the traditional distillation technique, defensive distillation aims to smooth the model during the training process by generalizing examples outside the training data. The specific training steps are shown in Fig. \ref{Fig13}.
\begin{enumerate}
  \item [i)] The probability vectors produced by the first DNN are used to label the dataset. These new labels are called soft labels as opposed to hard class labels.
\end{enumerate}
\begin{enumerate}
  \item [ii)] The dataset to train the second DNN model can be the newly labeled data or a combination of hard and soft labels. Since the second model combines the knowledge of the first model, it has smaller size, better robustness and smaller computational complexity.
\end{enumerate}

The basic idea of defensive distillation is to generate smooth classifiers that are more resilient to AEs, reducing the sensitivity of the DNN to the input perturbation.  In addition, defensive distillation improves the generalization ability because it does not modify the neural network architecture. Therefore, it has low training overhead and no testing overhead. Although attack methods \cite{Carlini2016}, \cite{Carlini2017a} have demonstrated that the defensive distillation does not improve the robustness of neural networks significantly, the following three methods are still a good research direction for defending AEs:
\begin{enumerate}
\item[(i)] Consider defensive distillation under different types of the perturbation (FGSM, L-BFGS, etc.);
\end{enumerate}

\begin{enumerate}
\item[(ii)] Investigate the effect of distillation on other DNN models and AE constructing algorithms;
\end{enumerate}

\begin{enumerate}
\item[(iii)] Study various distance metrics such as $L_0, L_2, L_{\infty}$ between the original examples and the AEs.
\end{enumerate}

\begin{enumerate}
  \item [3)]{\emph{Detector}}
\end{enumerate}

\begin{figure}
\centerline{{\includegraphics[width=\linewidth]{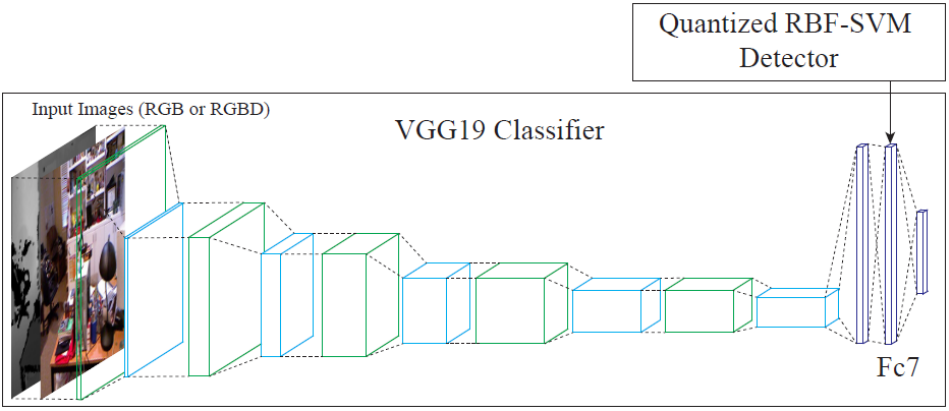}}}
\caption{\textbf{SafeNet architecture.} SafetyNet consists of a conventional classifier with an RBF-SVM that uses discrete codes to detect AEs \cite{Lu2017a}. }
\label{Fig14}
\end{figure}

Adversarial training is proposed to enhance the robustness of the model. However, this method lacks generalization ability and is difficult to popularize. Defensive distillation is proposed to defend AEs but defeated by a strong CW Attack. In 2017, Lu \emph{et al.} \cite{Lu2017a} proposed a radial basis function (RBF)-support vector machine (SVM)-based detector to detect whether the input is normal or adversarial (as shown in Fig. \ref{Fig14}). The detector can get the internal state of some back layers in the original classification neural network. If the detector finds out that the example is an AE, then it will be rejected.

We assume that the detector is difficult to be attacked, and the output of ReLU activation function is processed in the binary format. Since normal examples and AEs generate different binary codes, detectors can compare the code during the test to determine whether the input is normal or adversarial.

At present, AE detectors are mainly divided into the following classes \cite{Carlini2017}:

\textbf{Detection Based on Secondary Classification}. Generally, there are two kinds of secondary classification detection methods. One is the adversarial training detector \cite{Grosse2017}, \cite{Gong2017}, which is similar to adversarial training. The main idea is to add a new classification label to AEs during training. If an AE is detected, the model will classify it into a new class. Another is to take the characteristics extracted from AEs and original examples during the convolution layer as input. Then the labeled input data is used to train neural network detectors. This method performed well on detecting over 85$\%$ of AEs.

\textbf{Detection Based on Principal Component Analysis (PCA)}. The essence of PCA is to transform the original features linearly and map them to a low-dimensional space with the best possible representation of the original features. PCA-based detection methods are mainly divided into two types. The first one uses PCA in the input layer due to the greater weight of AEs processed by PCA than original examples \cite{Hendrycks2016}. The second one uses PCA in the hidden layer \cite{Li2017}. If the result of each hidden layer matches the feature of original examples, the detector will classify the input as original examples.

%
%
%
%

\begin{table*}
\caption{The advantages and disadvantages of defenses}
\label{Table5}
\centering
\begin{tabular}{|c|c|c|c|}
\hline
  Type & Method & Advantages & Disadvantages  \\
  \hline
  \multirow{3}{*}{Modified training/input} & Adversarial training \cite{Goodfellow2014}&
  \multicolumn{1}{c|}{\multirow{3}{*}{Simple, good defensive ability}}  &
  \multicolumn{1}{c|}{\multirow{3}{*}{Difficult to converge, high overhead
}} \\
  \cline{2-2}
   & Data compression \cite{Dziugaite2016} & \multicolumn{1}{c|}{}& \multicolumn{1}{c|}{} \\
   \cline{2-2}
   & Data randomization \cite{Xie2017}& \multicolumn{1}{c|}{}& \multicolumn{1}{c|}{}\\

   \hline
  \multirow{3}{*}{Modifying the network} & Deep Contractive Networks \cite{Gu2014}&
  \multicolumn{1}{c|}{\multirow{3}{*}{Low overhead, good generalization
}}  &
  \multicolumn{1}{c|}{\multirow{3}{*}{Model-dependent, high complexity
}} \\
  \cline{2-2}
   & Gradient masking \cite{Ross2017} & \multicolumn{1}{c|}{}& \multicolumn{1}{c|}{} \\
   \cline{2-2}
   & Defensive distillation  \cite{Papernot2016}& \multicolumn{1}{c|}{}& \multicolumn{1}{c|}{}\\

   \hline
  \multirow{2}{*}{Network add-ons} & GAN-based \cite{Lee2017}&
  \multicolumn{1}{c|}{\multirow{2}{*}{Low complexity, model-independent
}}  &
  \multicolumn{1}{c|}{\multirow{2}{*}{Weak generalization, not improving the robustness}} \\
  \cline{2-2}
   & Detection \cite{Lu2017a}& \multicolumn{1}{c|}{}& \multicolumn{1}{c|}{} \\

  \hline
\end{tabular}

\end{table*}

\textbf{Detection Based on Distribution}. There are two main distribution-based detection methods. The first one uses the maximum mean discrepancy \cite{Grosse2017} which measures the distance between two different but related distributions. Assuming that there are two sets of images $S_1$ and $S_2$, $S_1$ contains all the original examples, $S_2$ contains either all AEs or all original examples. If $S_1$ and $S_2$ have the same distribution, then $S_2$ has original examples; otherwise, $S_2$ is full of AEs. The second one uses kernel density estimation \cite{Feinman2017}. Since AEs have a different density distribution from the original examples, they can be detected with high confidence by the estimation of the density ratio. If the density ratio of one example is close to 1, it belongs to the original example. If the density ratio is much larger than 1, it belongs to AEs.

\textbf{Other Detection Methods}. Dropout randomization \cite{Hinton2014} is a method to use dropout randomly during AE detection. Original examples always generate correct labels, but AEs are of high possibility to be different from the label corresponding to original examples. In addition, another method called Mean Blur \cite{Li2017} uses the filter to perform mean blurring on the input image and can effectively improve the robustness of models.

\subsection{Other Defense Techniques }

Defenders' goal is to train a model where no AEs exist or AEs cannot be easily generated. Recently, some novel researches on defending AEs have been proposed.
Meng \emph{et al.} \cite{Meng2017} proposed a framework MagNet, including one or more separate detector networks and one reformer network. The detector network learns to distinguish normal examples from AEs by approximating normal examples. The reformer network moves AEs towards the normal examples. As MagNet is independent of the process of constructing AEs, hence it is effective in the black-box and gray-box attacks.
Dong \emph{et al.} \cite{Dong} proposed a high-level representation guided denoiser method which requires fewer training images and consumes less training time than previous defense methods at the expense of a reduced success rate.
Ma \emph{et al.} \cite{Ma2017} proposed Local Intrinsic Dimensionality (LID) to describe the dimensional attributes of the adversarial subspace in the AEs and proved that these features can distinguish normal examples from AEs effectively.
Baluja \emph{et al.} \cite{Baluja2017} proposed adversarial transformation networks (ATNs) to increase the diversity of perturbations to improve the effectiveness of adversarial training. However, ATNs may produce similar perturbations to iterative methods which are not suitable for adversarial training. In addition, hardware security primitives such as physical unclonable functions \cite{Zhang2014,Zhang2015,Zhang2019,Zhang2019b} can be used to randomize the model to assist the AE detection.

\subsection{Limitations of Defenses}

As discussed above, a lot of defenses have been proposed. In what follows, we summarize their advantages and disadvantages.

As shown in Table \ref{Table5}, adversarial training is simple and can significantly improve the robustness of models. However, AEs are required in the training process, which brings high overhead. Besides, it is difficult to theoretically explain which attack method to construct AEs for adversarial training can achieve the best robustness of models. Defensive distillation can greatly reduce the sensitivity to perturbations without modifying the neural network architectures. Therefore, defensive distillation incurs low overhead in training and testing. However, defensive distillation needs to add distillation temperature and modify the objective function, which increases the complexity of designing defensive models. Besides, attackers can easily bypass the defensive distillation by the following three strategies: 1) choose a more suitable objective function; 2) calculate the final layer of gradient instead of the second-to-last layer of gradient; 3) attack a fragile model and then transfer to the distillation model. Detectors do not need to modify the model architecture and parameters, hence the complexity is low. However, its performance is highly correlated with the type of detector. In addition, this method only detects the existence of AEs and does not improve the robustness of the model.

\section{Recent Challenges and New Opportunities}

AE construction and defense are one of the research hotspots in the AI security field. Although many AE construction methods and defense techniques have been proposed, various unresolved problems still exist. This section summarizes the challenges to this field and put forward to some future research directions.

In term of AE construction, there are three major challenges:

1) \textbf{It is difficult to build a generalized AE construction method.} In recent years, a lot of AE-construction methods have been proposed, such as the gradient-based FGSM, JSMA, the decision boundary-based DeepFool and the ensemble attack method combining multiple models. These methods are difficult to construct a generalized AE and can only achieve good performance in some evaluation metrics. Therefore, defenders can propose efficient defenses against these specific attacks. For example, the gradient can be hidden or obfuscated to prevent against the gradient-based AE construction methods.


2) \textbf{It is difficult to control the magnitude of perturbation for target images.} In the mainstream attack methods, attackers construct AEs by perturbing target images to fool neural network models. However, it is difficult to control the magnitude of perturbations because too small perturbations can not generate AEs and too large perturbations can be perceived by human eyes easily.

3) \textbf{AEs are difficult to maintain adversarial stability in real-world applications.} The image perturbed at specific distances and angles may result in the misclassification of the model. However, a lot of images perturbed at different distances and angles fail to fool the classifier \cite{Lu2017}. Moreover, AEs may lose their adversarial with physical transformation such as blurring, rotation, scaling and illumination \cite{Athalye2017}. Actually, it is hard for AEs to maintain stability in real-world applications.

Therefore, to address these issues, we propose to improve AE quality in the following three directions.

1) \textbf{Construct AEs with a high transfer rate.} With the diversification of neural network models, the effectiveness of attacks for a single model is not enough. Based on the transferability, constructing AEs with a high transfer rate is a prerequisite to evaluate the effectiveness of black-box attacks and a key metric to evaluate generalized attacks.

2) \textbf{Construct AEs without perturbing the target image.} When constructing AEs, the magnitude of perturbations to the target image is determined by experiments. Hence, the optimal perturbation will be different in various models. It increases the complexity of attacks and affects the success rate and the transfer rate. Therefore, constructing AEs without perturbing the target image is a novel and challenging research direction.

3) \textbf{Model the physical transformation.} In the physical world, attackers need to consider not only the magnitude of the perturbations but also the physical transformations such as translation, rotation, brightness, and contrast. However, it is difficult for attackers to use traditional algorithms to generate real-world AEs with high adversarial stability. Therefore, modeling physical perturbations is an efficient way to improve the stability of real-world AEs.

In terms of defending against AEs, there are two main challenges at present.

1) \textbf{Defense is highly related to model architectures and parameters.} The black-box attack does not need to obtain model architecture and parameters to construct AEs. Therefore, it is difficult for defenders to resist the black-box attack by modifying the model architectures or parameters. For example, defensive distillation needs to modify and retrain the target classifier.

2) \textbf{Weak generalization for defense models.} Adversarial training and detector are representative defense techniques. Adversarial training can improve the robustness of the model by adding AEs to the training set. The detector can detect examples based on AEs in the dataset. However, the defense effect is quite different when defending AEs generated by different attack methods, i.e., the generalization ability of defense models is weak.

\section{Conclusion}

DNNs have recently achieved state-of-the-art performance on a variety of pattern recognition tasks. However, recent researches show that DNNs, like many other ML models, are vulnerable to AEs. Although many AE construction and defense methods have been proposed, there are still some challenges to be solved. The state-of-the-art research is still in the adversarial development stage of ``while the priest climbs a post, the devil climbs ten''. In this survey, we review the state-of-the-art AE construction methods and the corresponding defense techniques, then summarize several challenges along with the future trends in this field. Although AEs have caused the deep learning to be questioned, it also prompts both academia and industry to understand the difference between AI and our human brain better.


\begin{IEEEbiography}[{\includegraphics[width=1in,height=1.25in,clip,keepaspectratio]{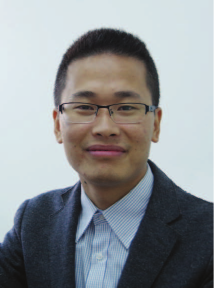}}]{Jiliang Zhang} (M'15-SM'19) received the Ph.D. degree in Computer Science and Technology from Hunan University, Changsha, China in 2015. From 2013 to 2014, he worked as a Research Scholar at the Maryland Embedded Systems and Hardware Security Lab, University of Maryland, College Park. From 2015 to 2017, he was an Associate Professor with Northeastern University, China. Since 2017, he has joined Hunan University. His current research interests include hardware/hardware-assisted security, artificial intelligence security, and emerging technologies.

Dr. Zhang was a recipient of the Hu-Xiang Youth Talent, and the best paper nominations in International Symposium on Quality Electronic Design 2017. He has been serving on the technical program committees of many international conferences such as ASP-DAC, FPT, GLSVLSI, ISQED and AsianHOST, and is a Guest Editor of the Journal of Information Security and Applications and Journal of Low Power Electronics and Applications.
\end{IEEEbiography}


\begin{IEEEbiography}[{\includegraphics[width=1in,height=1.25in,clip,keepaspectratio]{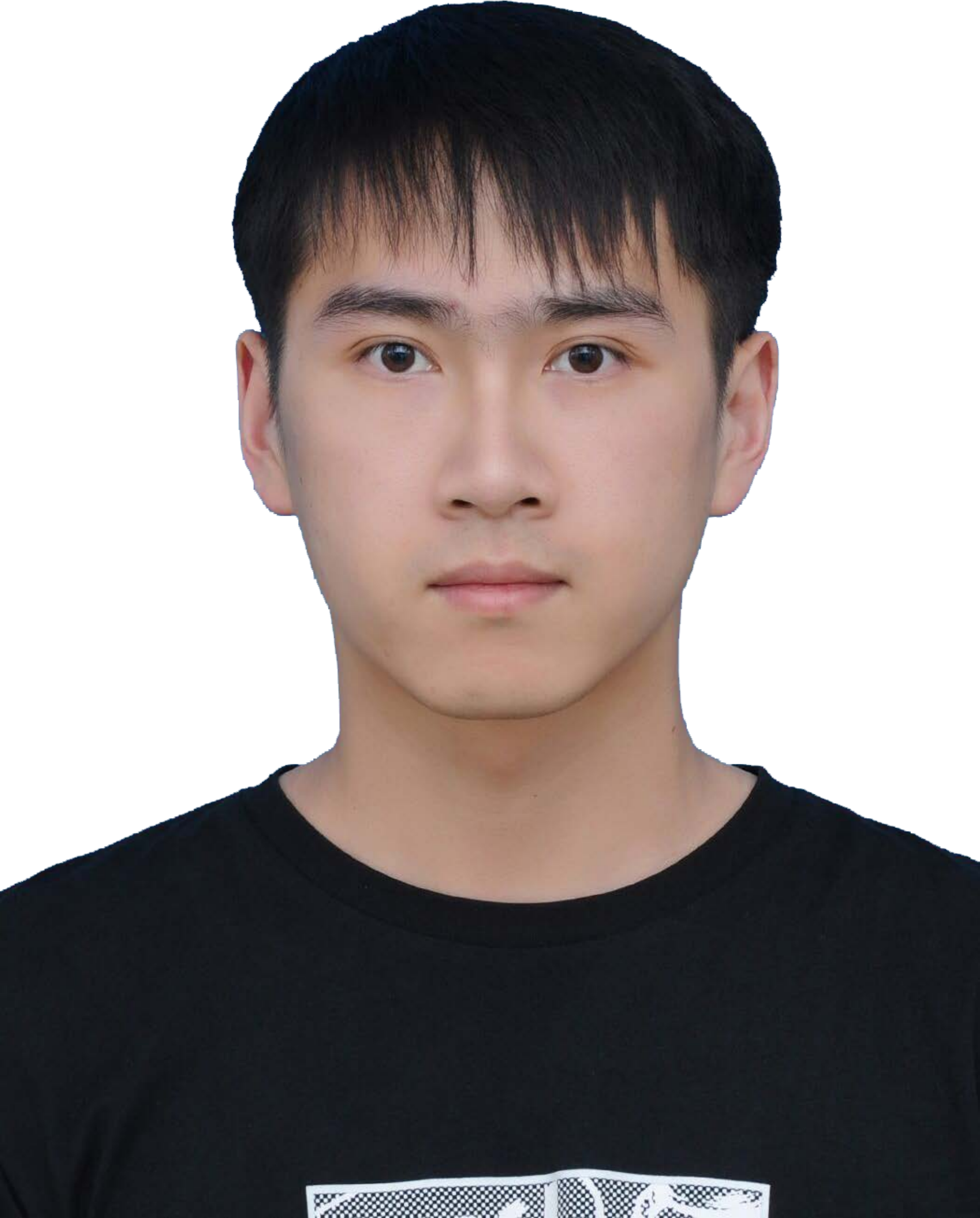}}]{Chen Li} received the B.S. degree in computer science and technology from Nanjing Audit University, Nanjing, China, in 2017. He is currently pursuing the M.S. degree in software engineering with Hunan University, Changsha, China.

His current research interests include artificial intelligence (AI) security and privacy.
\end{IEEEbiography}

\end{document}